\documentclass{article}

\usepackage{microtype}
\usepackage{graphicx}
\usepackage{subfigure}
\usepackage{booktabs} %

\usepackage{hyperref}

\usepackage[accepted]{icml2024}

\usepackage{amsmath}
\usepackage{amssymb}
\usepackage{mathtools}
\usepackage{amsthm}

\usepackage[capitalize,noabbrev]{cleveref}

\usepackage{bm}
\usepackage{xspace}
\newcommand{\methodname}{FOA\xspace}

\usepackage{threeparttable}
\usepackage{multicol,multirow}
\usepackage{balance}

\usepackage{colortbl}  %
\usepackage{xcolor}

\usepackage{hyperref}
\hypersetup{
    colorlinks=true,
    linkcolor=red,
    citecolor=teal,
    urlcolor=cyan,
}
\usepackage{enumitem}
\usepackage{pifont}
\let\oldding\ding%
\renewcommand{\ding}[2][1]{\scalebox{#1}{\oldding{#2}}}%
\usepackage{wrapfig}

\theoremstyle{plain}

\theoremstyle{definition}

\theoremstyle{remark}

\usepackage[textsize=tiny]{todonotes}

\usepackage{amsmath}

  \def\mC{{\mathcal C}}
  \def\mD{{\mathcal D}}

  \def\mL{{\mathcal L}}
  
  \def\mN{{\mathcal N}}

  \def\mX{{\mathcal X}}
  \def\mY{{\mathcal Y}}

  \DeclareMathAlphabet\mathbfcal{OMS}{cmsy}{b}{n}

  \def\0{{\bf 0}}
  \def\1{{\bf 1}}

  \def\bE{{\bf E}}

  \def\bI{{\bf I}}

  \def\bd{{\bf d}}
  \def\be{{\bf e}}

  \def\bmm{{\bf m}}

  \def\bp{{\bf p}}

  \def\bx{{\bf x}}
  \def\by{{\bf y}}

  \def\bx{{\bf x}}
  
  \def\by{{\bf y}}

  \def\bmm{{\bf m}}

  \def\bmu{{\bm \mu}}

  \def\bp{{\bf p}}

  \DeclareMathOperator*{\argmin}{arg\,min}
  
\def\eg{\emph{e.g.}} 
\def\ie{\emph{i.e.}} 
 
\def\etc{\emph{etc.}} \def\vs{\emph{vs.}}
\def\wrt{{w.r.t.~}}

\icmltitlerunning{Test-Time Model Adaptation with Only Forward Passes}

\begin{document}

\twocolumn[
\icmltitle{Test-Time Model Adaptation with Only Forward Passes}

\icmlsetsymbol{equal}{*}

\begin{icmlauthorlist}
\icmlauthor{Shuaicheng Niu}{ntu,webank}
\icmlauthor{Chunyan Miao}{ntu,webank,lily}
\icmlauthor{Guohao Chen}{tencent}
\icmlauthor{Pengcheng Wu}{ntu,webank}
\icmlauthor{Peilin Zhao}{tencent}
\end{icmlauthorlist}

\icmlaffiliation{ntu}{College of Computing and Data Science, Nanyang Technological University, Singapore}
\icmlaffiliation{webank}{Joint NTU-WeBank Research Centre on Fintech, Singapore}
\icmlaffiliation{lily}{
Joint NTU-UBC Research Centre of Excellence in Active Living for the Elderly (LILY), Singapore}
\icmlaffiliation{tencent}{Tencent AI Lab, Shenzhen, China}

\icmlcorrespondingauthor{Shuaicheng Niu}{shuaicheng.niu@ntu.edu.sg}

\icmlkeywords{Machine Learning, ICML}

\vskip 0.3in
]

\printAffiliationsAndNotice{}  %

\begin{abstract}
Test-time adaptation has proven effective in adapting a given trained model to unseen test samples with potential distribution shifts. However, in real-world scenarios, models are usually deployed on resource-limited devices, \eg, FPGAs, and are often quantized and hard-coded with non-modifiable parameters for acceleration. In light of this, existing methods are often infeasible since they heavily depend on computation-intensive backpropagation for model updating that may be not supported. To address this, we propose a test-time Forward-Optimization Adaptation (FOA) method. In FOA, we seek to solely learn a newly added prompt (as model's input) via a derivative-free covariance matrix adaptation evolution strategy. To make this strategy work stably under our online unsupervised setting, we devise a novel fitness function by measuring test-training statistic discrepancy and model prediction entropy. Moreover, we design an activation shifting scheme that directly tunes the model activations for shifted test samples, making them align with the source training domain, thereby further enhancing adaptation performance. Without using any backpropagation and altering model weights, FOA runs on quantized 8-bit ViT outperforms gradient-based TENT on full-precision 32-bit ViT, while achieving an up to \textit{24}-fold memory reduction on ImageNet-C. 
Code: \href{https://github.com/mr-eggplant/FOA}{https://github.com/mr-eggplant/FOA}.

\end{abstract}

\section{Introduction}
Deep neural networks often struggle to generalize when testing data encounter unseen corruptions or are drawn from novel environments~\cite{hendrycks2019benchmarking,koh2021wilds}, known as \textit{distribution shifts}. To address this, various methods have been extensively investigated in existing literature, such as domain generalization\cite{shankar2018generalizing,dou2019domain}, data augmentation~\cite{hendrycks2020augmix,yao2022improving} and unsupervised domain adaptation~\cite{saito2018maximum,zhang2020collaborative,Qiu2021CPGA}.

\begin{table}[t]
\vspace{-0.1in}
    \caption{Comparison \wrt prior gradient-based Test-Time Adaptation (TTA) \vs~our Forward-Optimization Adaptation. The memory usage and accuracy are measured via ViT-Base and batch size 64 on ImageNet-C (level 5). The memory of 8-bit ViT is an ideal estimation by 0.25\small{$\times$} memory of 32-bit ViT per~\citet{liu2021post}.}
    \label{tab:diff_between_foa_prior}
\newcommand{\tabincell}[2]{\begin{tabular}{@{}#1@{}}#2\end{tabular}}
 \begin{center}
 \begin{threeparttable}
    \resizebox{1.0\linewidth}{!}{
  \begin{tabular}{l|ll}
 	 \textit{~} & Prior TTA & Forward-Optimization Adaptation\\
    \cmidrule{1-3}       
         Update model weights & \ding{52} & \ding{55} \\
        \rowcolor{black!8} Backward propagation & \ding{52} & \ding{55} \\ 
         \multirow{3}{*}{Model compatibility} &  Full precision models  & Full precision models (32-bit) \\ 
          ~ & (32-bit) & Quantized models: \\
         & & ~~~$\bullet$ 8-bit, 6-bit, ...\\
        \rowcolor{black!8} & High-performance  & High-performance GPU   \\ 
        \rowcolor{black!8} Device & GPU & Low-power edge devices:\\
        \rowcolor{black!8} compatibility & ~ & ~~~$\bullet$ smartphones, iPads, FPGAs \\
        \rowcolor{black!8} ~ & ~ & ~~~$\bullet$ embodied robots, ... \\
        \multirow{2}{*}{Accuracy}  & 59.6\% (TENT, full & \textbf{66.3\% (full precision, 32-bit)} \\
         & precision, 32-bit) & \textbf{63.5\% (quantized, 8-bit)} \\
        \rowcolor{black!8}{Run-time} &  5,165 MB (TENT, full & \textbf{832 MB (full precision, 32-bit)}  \\ 
        \rowcolor{black!8} memory usage &   precision, 32-bit) & \textbf{208 MB (quantized, 8-bit)} \\
	\end{tabular}
	}
	 \end{threeparttable}
  \vspace{-0.2in}
	 \end{center}
\end{table}

Recently, test-time adaptation (TTA)~\cite{sun2020test,niu2023towards,iwasawa2021test,bartler2022mt3,liang2023comprehensive} has emerged as a rapidly progressing research area, with the aim of addressing domain shifts during test time.
By utilizing each data point once for immediate adaptation post-inference, TTA stands out with its minimal overhead compared to prior areas, making it well-suited for real-world applications.
According to whether involving backward propagation, existing TTA methods can generally be divided into the following two categories.

\textit{Gradient-free} methods learn from test data by adapting the statistics in batch normalization layers~\cite{schneider2020improving,khurana2021sita,lim2023ttn}, correcting the output probabilities~\cite{boudiaf2022parameter}, or adjusting the classifier~\cite{iwasawa2021test}, \etc~These methods, which avoid backpropagation and do not alter the original model weights, inherently reduce the risk of forgetting on source domain. However, their limited learning capacity, primarily stemming from the constraint of not explicitly exploiting the model feedback regarding given test samples to facilitate optimization with learnable parameters, may lead to suboptimal performance on out-of-distribution test data (as shown in Table~\ref{tab:imagenet-c-full-precision}). %
 
\textit{Gradient-based} methods~\cite{sun2020test,wang2021tent,goyal2022test} unleash the power of TTA by online updating model parameters through self-/un-supervised learning during testing. These methods encompass a variety of techniques including, but not limited to, rotation prediction~\cite{gidaris2018unsupervised}, contrastive learning~\cite{bartler2022mt3,liu2021ttt++}, entropy minimization~\cite{wang2021tent}, \etc~Although gradient-based TTA is effective in handling domain shifts, it still faces critical challenges when being deployed to real-world scenarios, as shown in Table~\ref{tab:diff_between_foa_prior}.

Firstly, deep models are usually deployed on various edge devices, \eg, smartphones, and embedded systems. Unlike high-performance GPUs, these devices typically possess limited computational power and memory capacity, insufficient for the intensive computations required by TTA, which often requires one or multiple rounds of backpropagation for each test sample~\cite{wang2021tent,zhang2021memo}.

Secondly, for resource or efficiency considerations,  deep models often undergo quantization before deployment – a process of reducing precision, \eg, from 32-bit to 8-bit. However, the non-differentiability of the discrete quantizer would result in vanishing gradients when propagated through multiple layers~\cite{Louizos2019relaxed}. This makes the deployed models incapable of supporting backpropagation operations, which are essential for prior TTA methods.

Lastly, on some specialized computational chips that are tailored for specific models~\cite{dass2023vitality,you2023vitcod}, the model parameters are often hard-coded and non-modifiable. This rigidity of model parameters poses another barrier to the implementation of TTA.

\begin{figure*}[t]
\centering
\includegraphics[width=1.0\linewidth]{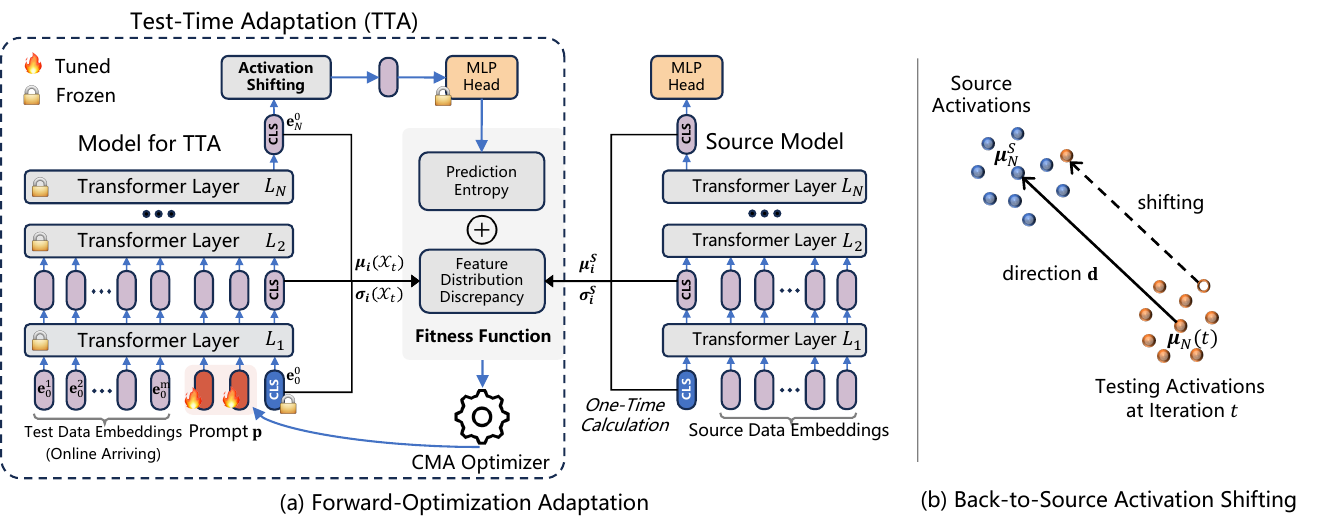}
\vspace{-0.3in}
\caption{(a) An illustration of our proposed \methodname. For each batch of online incoming test samples, we feed them alongside prompts $\bp$ into the TTA model, and calculate a fitness value that serves as a learning signal, aiding the covariance matrix adaptation (CMA) optimizer in learning the prompts $\bp$. This fitness function is derived from both the prediction entropy and the distribution discrepancy between the testing \texttt{CLS} activations and source \texttt{CLS} activations (calculated once). (b) We further boost the adaptation performance by directly adjusting the activations (before the final MLP head), guiding them from the testing distribution towards the source distribution.}
\label{fig:overall_illustration}
\end{figure*}

To address the above issues, we propose a test-time Forward-Optimization Adaptation (\methodname) method. Specifically, we seek to explore a backpropagation-free optimizer, called covariance matrix adaptation (CMA) evolution strategy~\cite{hansen2016cma}, for online test-time model adaptation. However, naively applying CMA in the TTA setting is infeasible, as it is hard for CMA to handle ultra-high dimensional optimization problems (\eg, deep model training) and it relies on supervised learning signals. Therefore, we propose to solely update a newly inserted prompt (as the model's input, shown in Figure~\ref{fig:overall_illustration}) at test time to reduce the dimension of solution space and meanwhile avoid altering model weights. Then, to make CMA work stably without supervised signals, we devise a novel unsupervised fitness function to evaluate candidate solutions, which comprises both model prediction entropy and the activation statistics discrepancy between out-of-distribution (OOD) testing samples and source in-distribution (ID) samples. Here, only a small number of ID samples is needed for source statistics estimation, \ie, 32 samples are sufficient for ImageNet (see Figure~\ref{fig:sensitivity} (c)). Moreover, to further boost adaptation performance, we devise a forward-only back-to-source activation shifting mechanism to directly adjust the activations of OOD testing samples, along with a dynamically updated shifting direction from the OOD testing domain to the ID source domain.

\textbf{Main Contributions.} 1) We introduce a novel and practical paradigm to TTA, termed forward-optimization adaptation. This paradigm operates without depending on backpropagation and avoids modification to the model weights, significantly broadening the real-world applicability of TTA across various contexts, including smartphones, FPGAs, and quantized models. 2) We achieve the goal of forward-only adaptation by prompt adaptation and activation shifting, where we design a new fitness function that guarantees stable prompt learning using a covariance matrix adaptation-based optimizer under the online unsupervised setting, and efficiently align the sample's activations in the testing domain with the source training domain. 3) Extensive experiments on four benchmarks and full precision/quantized models verify our effectiveness. Our method on 8-bit quantized ViT outperforms gradient-based TENT on full-precision 32-bit ViT, with up to \textit{24}-fold run-time memory reduction.

\section{Preliminary and Problem Statement}

We briefly revisit ViT and TTA in this section for the convenience of our method presentation and put \textbf{detailed related work discussions into Appendix~\ref{sec:related_work}} due to page limits.

\textbf{Vision Transformer (ViT)}~\cite{dosovitskiy2021an}.  In this paper, we focus mainly on transformer-based vision models that are widely used in practice and are also hardware-friendly. We first revisit ViT here for the presentation convenience of our method.
Formally, for a plain ViT $f_{\Theta}(\cdot)$ with $N$ layers, let $\bE_i=\{\be_i^j, j\in\mathbb{N}, 0\leq j \leq m\}$ be the patch embeddings as the input of the $(i+1)$-th layer $L_{i+1}$, where $m$ is the number of image patches and $\be_i^0$ denote an extra learnable \textit{classification token} (\texttt{[CLS]}) of the $i$-th layer $L_i$, the whole ViT is formulated as:
\begin{align}
    \bE_i &= L_i(\bE_{i-1}), ~~~ i=1,...,N \label{eq:vit_feature_extractor}\\
    \hat{\by} &= \texttt{Head}(\be_N^0). \label{eq:vit_head}
\end{align}

\textbf{Test-Time Adaptation (TTA)}~\cite{sun2020test,wang2021tent}. 
Let $f_{\Theta}(\cdot)$ be the model trained on labeled training dataset $\mD_{train} = \{(\bx_i,y_i)\}_{i=1}^{N}$ and $\bx_i \sim P\left( \bx \right)$. During testing, $f_{\Theta}(\cdot)$ shall perform well on in-distribution (ID) test samples drawn from $P\left( \bx \right)$. However, given a set of out-of-distribution (OOD) testing samples $\mD_{test} = \{\bx_j\}_{j=1}^{M} \sim Q \left( \bx \right)$ and $Q\left( \bx \right) \neq P\left( \bx \right)$, the prediction performance of $f_{\Theta}(\cdot)$ would decrease significantly. To address this, TTA methods often seek to update the model parameters by minimizing some unsupervised/self-supervised learning objective when encountering a testing sample:
\begin{equation}\label{eq:tta_formula}
    \min_{\Tilde{\Theta}} \mL(\bx;\Theta),~ \bx \sim Q \left( \bx \right),
\end{equation}
where $\Tilde{\Theta} \subseteq \Theta$ denotes the model parameters involved for updating. In general, the TTA objective $\mL(\cdot)$ can be formulated as rotation prediction~\cite{sun2020test}, contrastive learning~\cite{bartler2022mt3}, entropy minimization~\cite{wang2021tent,niu2023towards}, \etc %

\textbf{Problem Statement.}
In practical applications, deep models are frequently deployed on devices with limited resources, such as smartphones and embodied agents, and sometimes are even deployed after quantization or hard coding with non-modifiable parameters. These devices typically lack the capability for backward propagation, especially with large-size deep models. However, for existing TTA methods, such as SAR~\cite{niu2023towards} and MEMO~\cite{zhang2021memo}, performing TTA necessitates one or more rounds of backward computation for each test sample. This process is highly memory- and computation-intensive, hindering the broad application of TTA methods in real-world scenarios.

\section{Approach}

In this paper, we propose a novel test-time Forward-Optimization Adaptation (\methodname) method, which is also model updating-free, to boost the practicality of test-time adaptation in various real-world scenarios. From Figure~\ref{fig:overall_illustration}, \methodname conducts adaptation on both the input level and the output feature level.  \textit{1) Input level:} \methodname inserts a new prompt as the model's input, and then solely updates this prompt online for out-of-distribution (OOD) generalization, employing a derivative-free optimizer coupled with a specially designed unsupervised fitness function (c.f. Section~\ref{sec:fopa}).  \textit{2) Output feature level:} a back-to-source activation shifting strategy seeks to further boost adaptation, which directly refines the activation features of the final layer, by aligning them from the OOD domain back to the source in-distribution (ID) domain (c.f. Section~\ref{sec:back_to_source_activation_shifting}). 
We summarize the pseudo-code of \methodname in Algorithm~\ref{alg:overall}.

\subsection{Forward-Only Prompt Adaptation}\label{sec:fopa}

Unlike prior TTA methods that update model weights using backpropagation, we aim to achieve the goal of test-time out-of-distribution generalization in a backpropagation-free manner. To this end, we explore a derivative-free optimizer for TTA, namely \textbf{covariance matrix adaptation (CMA) evolution strategy}~\cite{hansen2016cma}. However, naively applying CMA to our TTA context is infeasible, the reasons are due to: 1) For TTA, the model parameters needing update are high-dimensional (even for methods like TENT~\cite{wang2021tent} that only updates the affine parameters of normalization layers), since the deep models are often with millions of parameters. This makes CMA intractable for direct deep model adaptation. 2) Conventional CMA methods rely on supervised offline learning, \ie, using ground-truth labels to assess candidate solutions. In contrast, TTA operates without ground-truth labels and typically in an online setting, rendering conventional CMA methods inapplicable. We empirically illustrated these issues in Table~\ref{tab:design_choices}.

To make CMA work in TTA, we introduce a new prompt as the model's input (as in Figure~\ref{fig:overall_illustration} (a)) for updating, thereby reducing the dimension of solution space and lowering the complexity for CMA optimization, meanwhile avoiding alter model weights. Then, we devise an unsupervised fitness function to provide consistent and reliable learning signals for the CMA optimization. We depict them in the following.

\begin{algorithm}[t]
	\caption{\textbf{F}orward-\textbf{O}ptimization \textbf{A}daptation (\methodname).}
	\label{alg:overall}
	\begin{algorithmic}[1]
    \INPUT{Batches of test samples $\{\mX_{t}\}_{t=1}^{T}$,  model $f_{\Theta}(\cdot)=\texttt{Head}(L_i(\cdot))$,  ID statistics $\{\bm\mu_i^S, {\bm\sigma}_i^S\}_{i=0}^N$, pop. size $K$.}
    \STATE Initialize $\bmm^{(0)}=\bf{0}, \bf{\Sigma}^{(0)}=\bI, \tau^{(0)}=1$ in Eqn. (\ref{eq:cma_es}).
    \FOR{$t=1, 2, ..., T$}
    \STATE Sampling $K$ prompt solutions $\{\bp_k^{t}\}_{k=1}^{K}$ by Eqn. (\ref{eq:cma_es}).
    \FOR{$k=1,...,K$}
    \STATE Calculate all layers' \texttt{CLS} features $\{\be_n^0\}_{n=1}^N$ using Eqn.~(\ref{eq:vit_feature_extractor}) with input $[\bp_k^{t}; \mX_t]$.
    \STATE Adjust $\be_N^0$ to source domain by Eqn.~(\ref{eq:activation shifting}).
    \STATE Predict $\hat{\mY}^k_t$ by \texttt{Head}$(\be_N^0)$.
    \STATE Calculate fitness value $v_k$ per Eqn.~(\ref{eq:fitness_function}).
    \ENDFOR
    \STATE Update $\bmm^{(t)}, \bf{\Sigma}^{(t)}, \tau^{(t)}$ according to $\{v_k\}_{k=1}^K$ using the CMA-ES algorithm~\cite{hansen2016cma}.
    \STATE Select final $\hat{\mY}_t$ from $\{\hat{\mY}_t^k\}_{k=1}^K$ with best $v_k$.
    \ENDFOR
    \OUTPUT The predictions $\{\hat{\mY}_t\}_{t=1}^T$.
	\end{algorithmic}
\end{algorithm}

\textbf{CMA-Based Prompt Adaptation.}
Inspired by the demonstrated effectiveness of continuous prompt learning in the field of deep model fine-tuning~\cite{jia2022visual,bahng2022exploring}, we add new prompt embeddings at the beginning of the model input (\ie, before the first transformer layer) for test-time updating, while keeping all other model parameters frozen. In this way, the dimension of learnable model parameters shall be significantly reduced and thus is compatible with CMA optimization. Formally, given a test sample $\bx\sim Q(\bx)$ and a ViT model $f_\Theta(\cdot)=\texttt{Head}(L_i(\cdot))$, our goal is to find an optimal prompt $\bp^*$:
\begin{equation}\label{eq:prompt_by_cma}
    \bp^*=\argmin_{\bp}\mL(f_{\Theta}(\bp;\bx)),
\end{equation}
where $\mL(\cdot)$ is a fitness function and $\bp\in\mathbb{R}^{d\times N_p}$ consists of $N_p$ prompt embeddings, each of dimension $d$. We solve this problem by employing the derivative-free CMA.

\textbf{Fitness Function for CMA.} 
To effectively solve Eqn.~(\ref{eq:prompt_by_cma}) using CMA, the primary challenge lies in developing an appropriate fitness $\mL(\cdot)$ to evaluate a given solution $\bp$. A direct approach might involve adopting existing TTA learning objectives, such as prediction entropy~\cite{wang2021tent}. However, this method encounters limitations when dealing with severely corrupted OOD samples, where model predictions are highly uncertain. In such cases, entropy-based measures struggle to provide consistent and reliable signals for CMA optimization. Moreover, focusing solely on optimizing entropy can lead the prompts towards degenerate and trivial solutions, as in Tables~\ref{tab:ablation_on_imageC} and~\ref{tab:design_choices}. To address these, we devise a new fitness to regularize the activation distribution statistics of OOD testing samples (forward propagated with optimized prompts), ensuring they are closely aligned with those from ID samples. This fitness functions at the distribution level, circumventing the issues of noise inherent in the uncertain predictions, thereby offering better stability.

\textit{Statistics calculation.} Before TTA, we first collect a small set of source in-distribution samples $\mD_S=\{\bx_q\}_{q=1}^{Q}$ and feed them into the model to obtain the corresponding \texttt{CLS} tokens $\{\be_i^0\}_{i=1}^N$. Then, we calculate the mean and standard deviations of \texttt{CLS} tokens $\{\be_i^0\}_{i=1}^N$ over all samples in $\mD_S$ to obtain source in-distribution statistics $\{\bm\mu_i^S, {\bm\sigma}_i^S\}_{i=0}^N$. Note that we only need a small number of in-distribution samples without labels for calculation, \eg, 32 samples are sufficient for the ImageNet dataset. Please refer to Figure~\ref{fig:sensitivity} (c) for the sensitivity analyses regarding this number. Similarly, we calculate the target testing statistics $\{\bm\mu_i(\mX_t)), {\bm\sigma}_i(\mX_t)\}_{i=0}^N$ over the current batch of testing samples $\mX_t$.

Based on the above, the overall fitness function for the $t$-th test batch samples $\mX_t$ is then given by:
\begin{align}\label{eq:fitness_function}
    \mL(f_{\Theta}(\bp;\mX_t)) &= \sum_{\bx\in\mX_t}\sum_{c\in\mC}-\hat{y}_c\log \hat{y}_c \nonumber \\ 
    + \lambda \sum_{i=1}^N &||\bmu_i(\mX_t)-\bmu_i^S||_2+||\bm\sigma_i(\mX_t)-\bm\sigma_i^S||_2,
\end{align}
where $\hat{y}_c$ is the $c$-th element of $\hat{\by}$ in Eqn.~(\ref{eq:vit_head}) \wrt sample $\bx$, and $\lambda$ is a trade-off parameter.

\textbf{CMA Evolution Strategy.} Instead of directly optimizing the prompt $\bp$ (in Eqn.~\ref{eq:prompt_by_cma}) itself, we learn a multivariate normal distribution of $\bp$ using a covariance matrix adaptation (CMA) evolution strategy~\cite{hansen2001completely,hansen2003reducing}. Here, we adopt CMA as it is one of the most successful and widely used evolutionary algorithms for non-convex black-box optimization in high-dimensional continuous solution spaces. To be specific, in each iteration $t$ (the $t$-th batch of test samples $\mX_t$), CMA samples a set/population of new candidate solutions/prompts (also known as individuals in evolution algorithms) from a parameterized multivariate normal distribution:
\begin{equation}\label{eq:cma_es}
    \bp_k^{(t)} \sim \bmm^{(t)} + \tau^{(t)}\mN(\bf{0}, \bf{\Sigma}^{(t)}).
\end{equation}
Here, $k\small{=}1,...,K$ and $K$ is the population size. $\bmm^{(t)}\small{\in}\mathbb{R}^{dN_p}$ is the mean vector of the search distribution at iteration step $t$, $\tau^{(t)}\small{\in}\mathbb{R}_+$ is the overall standard deviation that controls the step size, and $\bf{\Sigma}^{(t)}$ is the covariance matrix that determines the shape of distribution ellipsoid. Upon sampling the prompts $\{\bp_k^{(t)}\}_{k=1}^K$, we feed each $\bp_k^{(t)}$ along with the test sample $\mX_t$ into the model to yield a fitness value $v_k$ associated with $\bp_k^{(t)}$. Then, we update distribution parameters $\bmm^{(t)}$, $\tau^{(t)}$ and $\bf{\Sigma}^{(t)}$ based on the ranking of $\{v_k\}_{k=1}^K$ by maximizing the likelihood of previous candidate successful solutions (c.f.~\citet{hansen2016cma} for more algorithm details).

\subsection{Back-to-Source Activation Shifting}\label{sec:back_to_source_activation_shifting}

In this section, we propose a ``back-to-source activation shifting mechanism" to further boost the adaptation performance at the feature level, in cases of the above online prompt adaptation is inadequate. This shifting scheme directly alters the model's activations during inference and is notable for not requiring backpropagation. Specifically, given a test sample $\bx$, we move its corresponding $N$-th layer's \texttt{CLS} feature $\be_N^0$ (as shown in Eqn.~(\ref{eq:vit_head}), this feature is the input of the final task head), shifting them along the direction from out-of-distribution domain towards in-distribution domain:
\begin{align}
    \be_N^0 \leftarrow  \be_N^0 + \gamma \bd,
    \label{eq:activation shifting}
\end{align}
where $\bd$ is a shifting direction and $\gamma$ is a step size. We define $\bd$ as the vector extending from the center of out-of-distribution testing features to the center of source in-distribution features. In our online TTA setting, with the increase of testing samples, the center of testing features shall dynamically change. Thus, we update the shifting direction $\bd$ online by: 
\begin{equation}
    \bd_t = \bmu_N^S - \bmu_N(t),
    \label{eq:shifting direction}
\end{equation}
where $\bmu_N^S $ is the mean of the $N$-th final layer \texttt{CLS} feature $\be_N^0$ and calculated over source in-distribution samples $\mD_S$ (the same one used in Eqn.~(\ref{eq:fitness_function})). $\bmu_N(t)$ is the approximation of the overall test set statistics by exponential moving averages of statistics computed on sequentially arrived test samples. We define the mean estimate of the $\be_N^0$ in iteration $t$ (the $t$-th batch $\mX_t$) as:
\begin{align}
    \bmu_N(t) &= \alpha \bmu_N(\mX_t) + (1-\alpha)\bmu_{N}(t-1), \label{eq:testing_mu}
\end{align}
where $\bmu_N(\mX_t)$ is the mean of the $N$-th layer's \texttt{CLS} feature and calculated over the $t$-th test batch $\mX_t$. $\alpha\in[0,1]$ is a moving average factor and we set it to 0.1.

\section{Experiments}

\textbf{Datasets and Models.}  We conduct experiments on four benchmarks for OOD generalization, \ie, ImageNet-C~\cite{hendrycks2019benchmarking} (contains corrupted images in 15 types of 4 main categories and each type has 5 severity levels), ImageNet-R (various artistic renditions of 200 ImageNet classes)~\cite{hendrycks2021many}, ImageNet-V2~\cite{recht2019imagenet}, ImageNet-Sketch~\cite{wang2019learning}. We use ViT-Base~\cite{dosovitskiy2021an} as the source model for all experiments, including both full precision and quantized ViT models.
The models are trained on the source ImageNet-1K training set and the model weights are obtained from the \texttt{timm} repository~\cite{rw2019timm}. We adopt PTQ4ViT~\cite{yuan2022ptq4vit} for 8-bit and 6-bit model quantization. \textit{Unless stated otherwise, all ViT-Base models used in this paper are full precision with 32 bits.}

\textbf{Compared Methods.} We compare our proposed \methodname with two categories of TTA methods. 1) \textit{Gradient-free} methods: LAME~\cite{boudiaf2022parameter} is a post-training adaptation method by adjusting the model's output probabilities; T3A~\cite{iwasawa2021test} updates a prototype-based classifier during test time. 2) \textit{Gradient-based} methods: TENT~\cite{wang2021tent} optimizes the affine parameters of norm layers by minimizing the prediction entropy of test samples and SAR~\cite{niu2023towards} further optimizes the prediction entropy via active sample selection and a sharpness-aware optimizer; CoTTA~\cite{wang2022continual} adapts a given model via augmentation-based consistency maximization and a teacher-student learning scheme.

\textbf{Implementation Details.} We set the number of prompt embeddings $N_p$ to 3 and initialize prompts with uniform initialization. We set the batch size ($BS$) to 64 by following TENT and SAR for fair comparisons. The population size $K$ is set to $28=4+3\times\log(prompt ~~dim)$ by following~\cite{hansen2016cma} and $\lambda$ in Eqn.~(\ref{eq:fitness_function}) is set to 0.4$\times BS/64$ on ImageNet-C/V2/Sketch, and 0.2$\times BS/64$ on ImageNet-R to balance the magnitude of two losses. We use the validation set of ImageNet-1K to estimate source ID statistics.  The step size $\gamma$ in Eqn.~(\ref{eq:activation shifting}) is set to 1.0, aiming to exactly align the overall center of testing and training features. The effect of each hyperparameter is investigated in Section~\ref{sec:ablation} and Appendix~\ref{suppl:more_ablations}. More 
implementation details of compared methods are put in Appendix~\ref{suppl:subsec:experiment protocols}. 

\textbf{Evaluation Metrics.} 1) Classification \textbf{Accuracy (\%, $\uparrow$)} on OOD testing samples, \ie, ImageNet-C/R/V2/Sketch. 2) \textbf{Expected Calibration Error (ECE) (\%, $\downarrow$)} measures the difference between predicted probabilities and actual outcomes in a probabilistic model~\cite{naeini2015ece}. ECE is important to evaluate the trustworthiness of model predictions, such as in medical diagnostics and auto driving.

\begin{table*}[t]
    \caption{Comparisons with SOTA methods on ImageNet-C (severity level 5) with ViT regarding \textbf{Accuracy (\%)}.  \textbf{BP} is short for \textbf{backward propagation} and the \textbf{bold} number indicates the best result. We only report average ECE (\%,$\downarrow$) here and put detailed ECEs in Appendix~\ref{suppl:more_exps}. }
    \label{tab:imagenet-c-full-precision}
\newcommand{\tabincell}[2]{\begin{tabular}{@{}#1@{}}#2\end{tabular}}
 \begin{center}
 \begin{threeparttable}
 \LARGE
    \resizebox{1.0\linewidth}{!}{
  \begin{tabular}{lcccccccccccccccc>{\columncolor{black!8}}c>{\columncolor{black!8}}c}
 	\multicolumn{1}{c}{} & \multicolumn{1}{c}{}& \multicolumn{3}{c}{Noise} & \multicolumn{4}{c}{Blur} & \multicolumn{4}{c}{Weather} & \multicolumn{4}{c}{Digital} & \multicolumn{2}{c}{Average} \\
 	 Method & BP & Gauss. & Shot & Impul. & Defoc. & Glass & Motion & Zoom & Snow & Frost & Fog & Brit. & Contr. & Elas. & Pix. & JPEG & Acc. & ECE \\
    \cmidrule{1-19}       
        NoAdapt & \ding{55} & 56.8  & 56.8  & 57.5  & 46.9  & 35.6  & 53.1  & 44.8  & 62.2  & 62.5  & 65.7  & 77.7  & 32.6  & 46.0  & 67.0  & 67.6  & 55.5  & 10.5  \\ 
        LAME & \ding{55} & 56.5  & 56.5  & 57.2  & 46.4  & 34.7  & 52.7  & 44.2  & 58.4  & 61.5  & 63.1  & 77.4  & 24.7  & 44.6  & 66.6  & 67.2  & 54.1 & 11.0 \\
        T3A & \ding{55} & 56.4  & 56.9  & 57.3  & 47.9  & 37.8  & 54.3  & 46.9  & 63.6  & 60.8  & 68.5  & 78.1  & 38.3  & 50.0  & 67.6  & 69.1  & 56.9  & 26.8  \\ 
TENT & \ding{52} & 60.3  & 61.6  & 61.8  & 59.2  & 56.5  & 63.5  & 59.2  & 54.3  & 64.5  & 2.3  & 79.1  & 67.4  & 61.5  & 72.5  & 70.6  & 59.6  & 18.5  \\   
CoTTA & \ding{52} & 63.6 & 63.8 & 64.1 & 55.5 & 51.1 & 63.6 & 55.5 & 70.0 & 69.4 & 71.5 & 78.5 & 9.7 & 64.5 & 73.4 & 71.2 & 61.7 & 6.5\\
SAR  & \ding{52}    & 59.2  & 60.5  & 60.7  & 57.5  & 55.6  & 61.8  & 57.6  & 65.9  & 63.5  & 69.1  & 78.7  & 45.7  & 62.4  & 71.9  & 70.3  & 62.7  & 7.0  \\ 
\cmidrule{1-19}
\methodname (ours)  & \ding{55}    & 61.5  & 63.2  & 63.3  & 59.3  & 56.7  & 61.4  & 57.7  & 69.4  & 69.6  & 73.4  & 81.1  & 67.7  & 62.7  & 73.9  & 73.0  & \textbf{66.3}  & \textbf{3.2 }\\ 
	\end{tabular}
	}
	 \end{threeparttable}
\vspace{-0.15in}	 
  \end{center}
\end{table*}

\begin{table}[t]
    \caption{Comparisons with state-of-the-art methods on ImageNet-R/V2/Sketch with ViT-Base.  \textbf{BP} is short for \textbf{backward propagation} and the \textbf{bold} number indicates the best result.}
    \label{tab:imagenet-rv2asketch-full-precision}
\newcommand{\tabincell}[2]{\begin{tabular}{@{}#1@{}}#2\end{tabular}}
 \begin{center}
 \begin{threeparttable}
 \LARGE
    \resizebox{1.0\linewidth}{!}{
  \begin{tabular}{lc|ccc>{\columncolor{black!8}}c|ccc>{\columncolor{black!8}}c}
 	\multicolumn{1}{c}{} & \multicolumn{1}{c}{}& \multicolumn{4}{c}{Accuracy (\%, $\uparrow$)} & \multicolumn{4}{c}{ECE (\%, $\downarrow$)} \\
 	 Method & BP & R & V2  & Sketch & Avg. & R & V2  & Sketch & Avg. \\
    \cmidrule{1-10}       
        NoAdapt & \ding{55} & 59.5  & 75.4    & 44.9 & 59.9 & 2.5  & 5.6    & 7.9 & 5.3 \\
        LAME & \ding{55} & 59.0 & 75.2  & 44.4 & 59.6 & 2.5 & 5.0  & 9.7 & 5.7\\
        T3A & \ding{55} & 58.0  & 75.5   & 48.5 & 60.7 & 25.9  & 23.4    & 37.4 & 28.9 \\
        TENT & \ding{52} & 63.9  & 75.2    & 49.1 & 62.7 & 7.2  & 4.5    & 22.8 & 11.5 \\ 
        CoTTA & \ding{52} & 63.5  & 75.4   & 50.0 & 62.9 & 2.8  & 3.4    & 17.9 & 8.0 \\
        SAR  & \ding{52}    & 63.3  & 75.1    & 48.7 & 62.4 & 3.0  & 2.7    & 16.5  & 7.4\\ 
\cmidrule{1-10}
\methodname (ours)  & \ding{55}    & 63.8  & 75.4    & 49.9 & \textbf{63.0} & 2.7  & 3.2    & 7.8 & \textbf{4.6}\\  
	\end{tabular}
	}
	 \end{threeparttable}
	 \end{center}
    \vspace{-0.15in}
\end{table}

\subsection{Results on Full Precision Models}

In this section, we compare our \methodname with existing state-of-the-art TTA methods on the full precision ViT-Base model. From the results on ImageNet-C in Table~\ref{tab:imagenet-c-full-precision}, we have the following observations. 1) Our \methodname achieves the best average accuracy and ECE over 15 different corruption types, suggesting our effectiveness. 2) Compared with NoAdapt, gradient-free methods LAME and T3A obtain slight performance gains or perform even worse, as they do not update core model weights and thus may suffer from limited learning capacity. Here, LAME performs worse than NoAdapt, because it only adjusts the model output logits and is not very effective when the OOD test sample stream does not suffer from prior label shifts, which is consistent with the results reported by LAME itself. 3) Compared with LAME and T3A, gradient-based methods (TENT, CoTTA and SAR) explicitly modify model parameters by optimizing unsupervised/self-supervised losses, and thus achieve much better performance, \eg, the average accuracy of 56.9\% (T3A) \vs~62.7\% (SAR). 4) Without using any back-propagation, our \methodname outperforms gradient-based SAR with 3.6\% average accuracy and 3.8\% average ECE gains, demonstrating our superiority in deploying to lightweight devices (\eg, smartphones and FPGAs) and quantized models. 5) \methodname achieves much lower average ECE compared with BP-based methods, \eg, 18.5\% (TENT) \vs~3.2\% (\methodname). This mainly benefits from our activation discrepancy regularization (in Eqn.~(\ref{eq:fitness_function})), which alleviates the error accumulation issue of prior methods that may employ imprecise pseudo labels or entropy for learning.
At last, from the results on ImageNet-R/V2/Sketch in Table~\ref{tab:imagenet-rv2asketch-full-precision}, our \methodname achieves the best or comparable performance \wrt both accuracy and ECE, further suggesting our effectiveness.

\subsection{Results on Quantized Models}

\begin{table*}[t]
    \caption{Effectiveness of our \methodname on \textbf{Quantized ViT models}. We report the corruption \textbf{Accuracy (\%)} and average ECE (\%, $\downarrow$) on ImageNet-C (severity level 5). The \textbf{bold} number indicates the best result and see Appendix~\ref{suppl:more_exps} for the detailed ECEs of each corruption.}
    \label{tab:quantized_results}
\newcommand{\tabincell}[2]{\begin{tabular}{@{}#1@{}}#2\end{tabular}}
 \begin{center}
 \begin{threeparttable}
 \LARGE
    \resizebox{1.0\linewidth}{!}{
  \begin{tabular}{llccccccccccccccc>{\columncolor{black!8}}c>{\columncolor{black!8}}c}
 	\multicolumn{1}{c}{} & \multicolumn{1}{c}{}& \multicolumn{3}{c}{Noise} & \multicolumn{4}{c}{Blur} & \multicolumn{4}{c}{Weather} & \multicolumn{4}{c}{Digital} & \multicolumn{2}{c}{Average} \\
 	 Model & Method & Gauss. & Shot & Impul. & Defoc. & Glass & Motion & Zoom & Snow & Frost & Fog & Brit. & Contr. & Elas. & Pix. & JPEG & Acc. & ECE  \\
    \cmidrule{1-19}       
        \multirow{3}{*}{8-bit} & NoAdapt  &         55.8  & 55.8  & 56.5  & 46.7  & 34.7  & 52.1  & 42.5  & 60.8  & 61.4  & 66.7  & 76.9  & 24.6  & 44.7  & 65.8  & 66.7  & 54.1  & 10.8 \\  
        & T3A  & 55.6  & 55.7  & 55.7  & 45.8  & 34.4  & 51.1  & 41.2  & 59.5  & 61.9  & 66.8  & 76.4  & 45.5  & 43.4  & 65.6  & 67.5  & 55.1  & 25.9 \\  
 &  \methodname (ours)  & 60.7  & 61.4  & 61.3  & 57.2  & 51.5  & 59.4  & 51.3  & 68.0  & 67.3  & 72.4  & 80.3  & 63.2  & 57.0  & 72.0  & 69.8  & \textbf{63.5}  & \textbf{3.8} \\   
\cmidrule{1-19}
\multirow{3}{*}{6-bit} &  NoAdapt  & 44.2  & 42.0  & 44.8  & 39.8  & 28.9  & 43.4  & 34.7  & 53.2  & 59.8  & 59.0  & 75.1  & 27.4  & 39.0  & 59.1  & 65.3  & 47.7  & 9.9 \\ 
& T3A     & 43.3  & 41.3  & 42.7  & 29.1  & 23.4  & 38.9  & 30.0  & 49.4  & 58.3  & 60.2  & 73.8  & 31.0  & 36.3  & 58.0  & 65.2  & 45.4  & 30.1 \\ 
 & \methodname (ours)    & 53.2  & 51.8  & 54.6  & 49.6  & 38.8  & 51.0  & 44.8  & 60.3  & 65.0  & 68.8  & 76.7  & 39.5  & 46.6  & 67.3  & 68.6  & \textbf{55.8} & \textbf{5.5} \\ 
	\end{tabular}
	}
 \vspace{-0.15in}
	 \end{threeparttable}
	 \end{center}
\end{table*}

In practical applications, deep models are often deployed on edge devices with efficiency considerations, undergoing a process known as quantization. These devices, constrained by limited resources, typically do not support backward propagation due to its high memory and computational demands. Consequently, traditional gradient-based TTA methods like TENT, CoTTA, and SAR are not viable in such settings. On the contrary, our \methodname is adaptable to these quantized models. We demonstrate this by applying \methodname to quantized ViT models and benchmarking it against T3A. As indicated in Table~\ref{tab:quantized_results}, \methodname outperforms T3A significantly in terms of both accuracy and ECE on 8-bit and 6-bit models. Notably, \methodname with an 8-bit ViT surpasses the performance of the gradient-based TENT method using a full precision 32-bit ViT on ImageNet-C, achieving 63.5\% accuracy (our \methodname, 8-bit) \vs~59.6\% (TENT, 32-bit). These results collectively underscore the superiority of our \methodname in such quantized model deployment scenarios.

\subsection{Ablation Studies}
\label{sec:ablation}

\textbf{Effectiveness of Components in \methodname.}
In our \methodname, we mentioned that naively applying CMA with entropy minimization in TTA is infeasible, and thus we propose an activation distribution discrepancy-based fitness function to guide the stable learning of CMA and an activation shifting scheme to boost the adaptation performance. We ablate them in Table~\ref{tab:ablation_on_imageC}. Firstly, CMA with \textit{Entropy} fitness performs poorer than ``NoAdapt", which verifies the necessity of devising a new fitness function. Secondly, our \textit{Activation Discrepancy} fitness works well to provide stable learning signals for CMA and improves the adaptation accuracy on ImageNet-C from 55.5\% to 63.4\%. Thirdly, even only with the \textit{Activation Shifting} scheme, it also improves the accuracy from 55.5\% to 59.1\%, suggesting its effectiveness. Lastly, by incorporating \textit{Entropy} and \textit{Activation Discrepancy} as a whole fitness function and with the \textit{Activation Shifting} scheme, our \methodname achieves the best performance, \ie, 66.3\% average accuracy and 3.2\% average ECE on ImageNet-C.

\begin{table}[t]
    \caption{Ablations of components in our \methodname. \textit{Entropy} and \textit{Activation (Act.) Discrepancy} are the left/right item in Fitness Function (Eqn. \ref{eq:fitness_function}) used for CMA-based prompt adaptation.  \textit{Act. Shifting} is the method proposed in Section~\ref{sec:back_to_source_activation_shifting}. We report the average results over 15 corruptions on ImageNet-C (level 5) with ViT-Base.}
    \label{tab:ablation_on_imageC}
\newcommand{\tabincell}[2]{\begin{tabular}{@{}#1@{}}#2\end{tabular}}
 \begin{center}
 \begin{threeparttable}
 \LARGE
    \resizebox{1.0\linewidth}{!}{
  \begin{tabular}{ccc|cc}
 	 \textit{Entropy} & \textit{Act. Discrepancy} & \textit{Act. Shifting} & Acc. (\%, $\uparrow$) & ECE (\%,$\downarrow$) \\
    \cmidrule{1-5}       
        \multicolumn{3}{c|}{NoAdapt} & 55.5  & 10.5  \\ 
        \ding{52} & ~ & ~ & 44.9  & 36.8  \\ 
        ~ & \ding{52} & ~ & 63.4  & 9.4  \\ 
        ~ & ~ & \ding{52} & 59.1  & 12.7  \\ 
        ~ & \ding{52} & \ding{52} & 63.8  & 9.9  \\ 
        \ding{52} & \ding{52} & ~ & 65.4  & 3.3  \\ 
    \cmidrule{1-5}
        \ding{52} & \ding{52} & \ding{52} & \textbf{66.3}  & \textbf{3.2} \\ 
	\end{tabular}
	}
	 \end{threeparttable}
	 \end{center}
  \vspace{-0.15in}
\end{table}

\begin{figure*}
  \centering
  \subfigure{\includegraphics[width=0.333\textwidth]{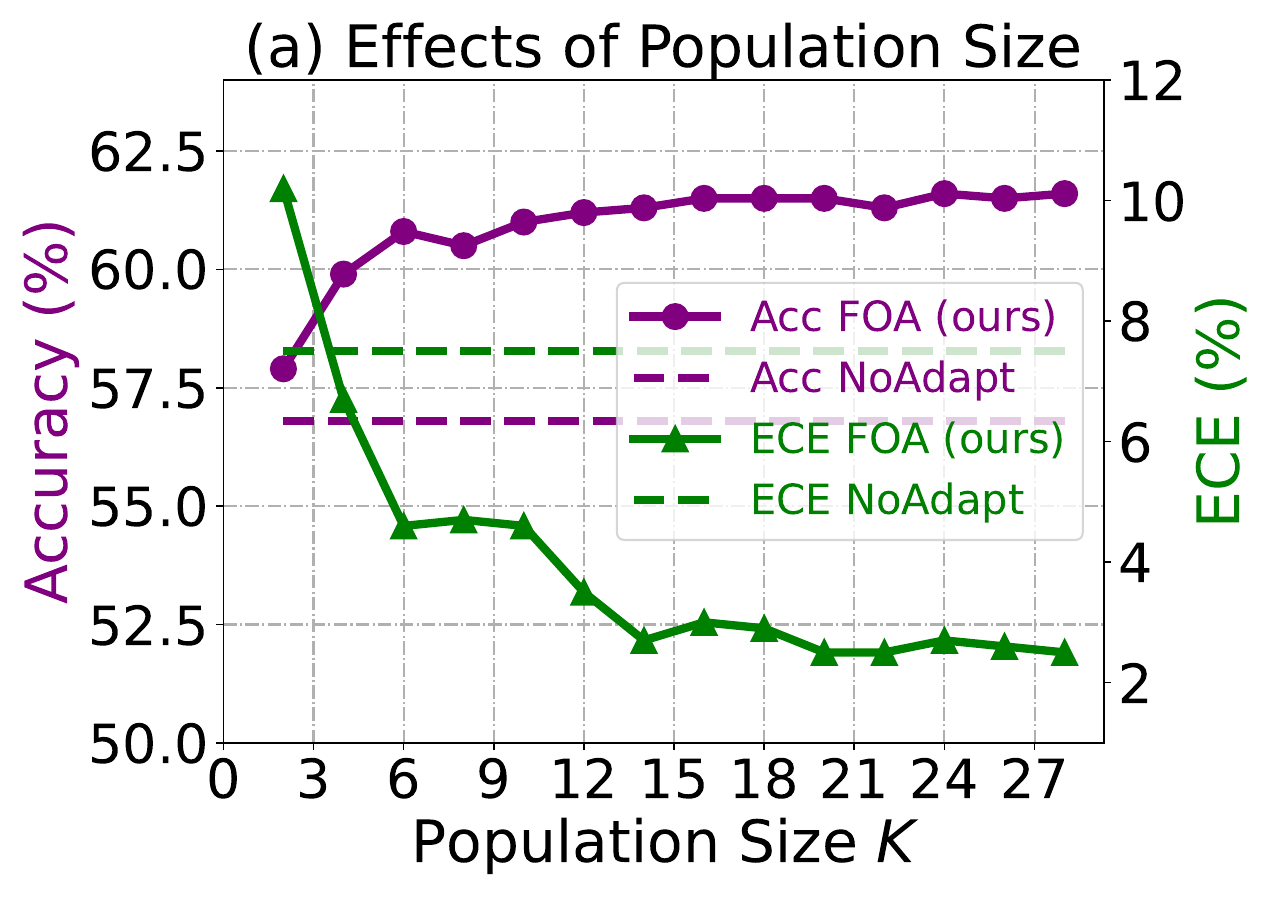}}\hspace{-0.1in}
  \subfigure{\includegraphics[width=0.333\textwidth]{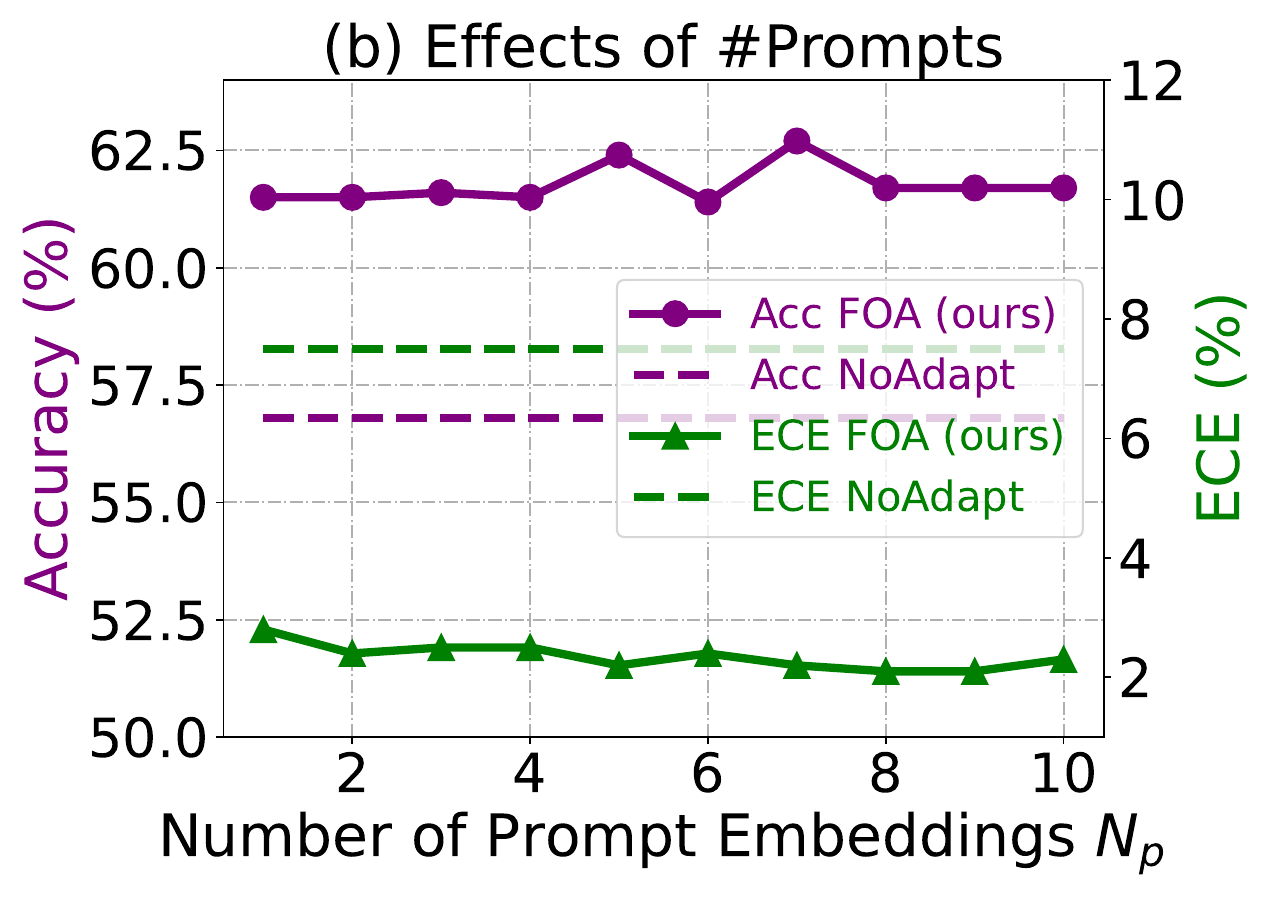}}\hspace{-0.1in}
  \subfigure{\includegraphics[width=0.333\textwidth]{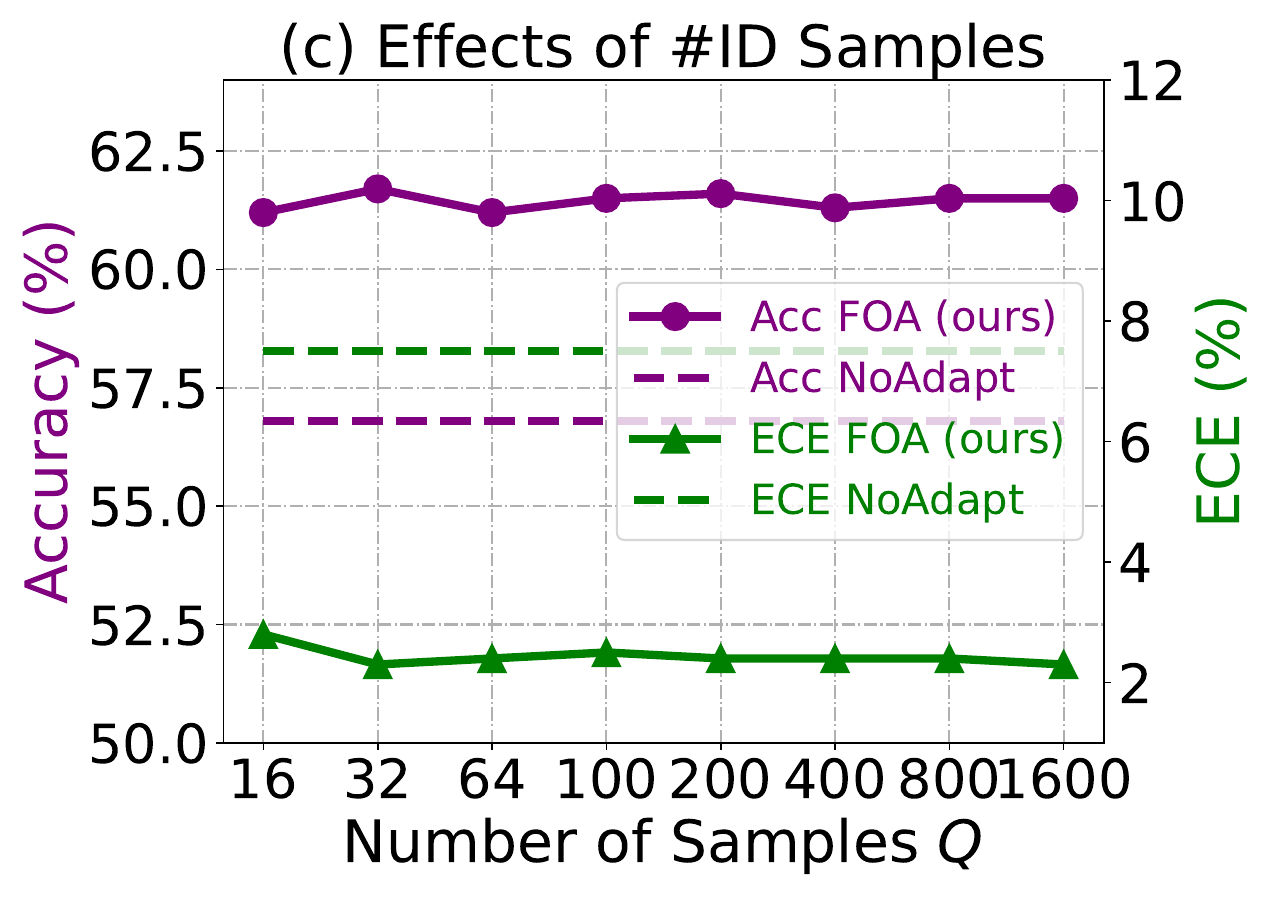}}
  \vspace{-0.12in}
  \caption{Parameter sensitivity analyses of our \methodname. Experiments are conducted on ImageNet-C (Gaussian Noise, level 5) with ViT-Base.}
  \label{fig:sensitivity}
  \vspace{-0.1in}
\end{figure*}

\textbf{Effects of Population Size $K$ in CMA (Eqn.~(\ref{eq:cma_es})).}
We evaluate our \methodname with different $K$ from \{2, 3, ..., 28\}. From Figure~\ref{fig:sensitivity} (a), the performance of our \methodname converges when $K\small{>}15$. 
Notably, at $K\small{=}2$, \methodname outperforms both NoAdapt and gradient-free T3A, \eg, 57.9\% (\methodname) \vs~56.4\% (T3A) regarding accuracy. And with $K\small{=}6$, \methodname surpasses the gradient-based TENT, \ie, 60.8\% accuracy (\methodname) \vs~60.3\% (TENT). These results demonstrate the effectiveness of \methodname under small $K$ values (the smaller $K$, the higher efficiency). Please refer to Table~\ref{tab:computation_complexity} for efficiency comparisons.

\textbf{Effects of Number of Prompt Embeddings $N_p$ in \methodname.}
In our \methodname, we add prompt embeddings in the input layer for CMA learning. Here, we evaluate \methodname with different numbers of $N_p$, selected from \{1, 2, ..., 10\}. In Figure~\ref{fig:sensitivity} (b), we observe that the performance of \methodname exhibits only minor variations across different $N_p$, showing a low sensitivity to $N_p$. Furthermore, setting $N_p$ to 5/7 results in marginally better performance, suggesting that this is a more optimal value. However, for all main experiments, we simply fix $N_p$ at 3 and do not carefully tune $N_p$, as access to testing data for parameter tuning is typically unavailable in practice.

\textbf{Effects of \#Samples ($Q$) for Calculating $\{\bm\mu_i^S, {\bm\sigma}_i^S\}_{i=0}^N$.} As described in Section~\ref{sec:fopa}, the calculation of source training statistics $\{\bm\mu_i^S, {\bm\sigma}_i^S\}_{i=0}^N$ involves a small set of unlabeled in-distribution samples,
which can be collected via existing OOD detection techniques~\cite{berger2021confidence} or directly using the training samples. Here, we investigate the effect
of \#samples needed, selected from \{16, 32, 64, 100, 200, 400, 800, 1600\}. From Figure~\ref{fig:sensitivity} (c), our \methodname consistently achieves
stable performance when \#samples greater than 32, regarding both the accuracy and ECE. These results
show that our \methodname does not need to collect too many in-distribution
samples, which are easy to obtain in practice.

\subsection{More Discussions}

\textbf{Results on Single Sample Adaptation (Batch Size = 1).}
In our \methodname, the prompt is learned over a batch of test samples each time. This process suffers from a challenge when the batch size is limited to one, as it requires the computation of the mean and variance of features, which may not be feasible with a single sample. Nonetheless, this issue is not significantly problematic in real-world applications. We propose a solution in the form of an interval update strategy, referred to as \textbf{\methodname-I}. Specifically, given an ongoing stream of test data, we opt to update the prompts (performing CMA optimization) after encountering a pre-defined number of samples, denoted as $I$. During this interval, we temporarily store the relevant features of all \texttt{CLS} tokens or the original image until the next update. From Table~\ref{tab:foa_on_single_sample_adaptation}, our \methodname-I is effective across various values of $I$. Notably, \methodname-I with an interval of $L\small{=}4$ outperforms the accuracy of TENT (with batch size 64), suggesting its effectiveness. Interestingly, \methodname-I with smaller intervals (\eg, $I\small{=}4$) shows better performance than it with $I\small{=}64$. This is because a smaller $I$ leads to more CMA iteration steps for a given test data stream.

\begin{table}[t]
    \caption{Effectiveness of \methodname with interval update strategy (different intervals $I$), termed \methodname-I, for single sample adaptation. We report results on ImageNet-C (Gaussian, level 5) with Vit-Base.}
    \label{tab:foa_on_single_sample_adaptation}
\newcommand{\tabincell}[2]{\begin{tabular}{@{}#1@{}}#2\end{tabular}}
 \begin{center}
 \begin{threeparttable}
 \LARGE
    \resizebox{1.0\linewidth}{!}{
  \begin{tabular}{lccccccc}
 ~& \multirow{2}{*}{\tabincell{c}{NoAdapt}} & TENT & \methodname-I  & \methodname-I  & \methodname-I  & \methodname-I  & \methodname-I  \\
  ~& ~ & ($BS\small{=}64$) &  ($I\small{=}4$) & ($I\small{=}8$) &  ($I\small{=}16$) &  ($I\small{=}32$) &  ($I\small{=}64$) \\
  \cmidrule{1-8}
      Acc. & 56.8 & 60.3 & 62.1 &	62.2 &	62.1 &	61.9 &	61.5  \\
      ECE & 7.5 & 13.7 &  3.3 &	2.6 &	2.8 &	2.7 &	2.5  \\
	\end{tabular}
	}
	 \end{threeparttable}
	 \end{center}
  \vspace{-0.1in}
\end{table}

\textbf{Run-Time Memory Usage.}
The memory usage during runtime is influenced by both the model and the batch size ($BS$). We detail the memory consumption of different methods with different $BS$ in Table~\ref{tab:run_time_peak_memory}. Our \methodname exhibits a marginally higher memory consumption compared to NoAdapt across different $BS$, due to the need to maintain certain feature statistics, \eg, \methodname requires 3MB extra memory than NoAdapt for $BS=4$. Notably, \methodname significantly lowers memory usage compared to existing gradient-based TTA, \eg, 832 MB (\methodname) \vs~5,165 MB (TENT) / 16,836 MB (CoTTA) under $BS=64$. Moreover, our \methodname-I V1/V2, which are variants (introduced in the last subsection) designed for single sample adaptation ($BS=1$), with different update intervals $I$ further decrease the memory footprint of \methodname. Lastly, applying \methodname to quantized models yields additional memory savings proportional to the quantization level. These results verify the efficiency of \methodname, particularly for deployment on resource-constrained edge devices.

\begin{table}[t]
    \caption{Comparison \wrt run-time memory (MB) usage. Results obtained via ViT-Base (32/8-bit) on ImageNet-C (Gaussian, level 5). \methodname-I V1/V2 denote storing features/images for interval update under batch size ($BS$) 1. The memory for 8-bit ViT is an ideal estimation by 0.25$\small{\times}$ memory of 32-bit ViT per~\citet{liu2021post}.}
    \label{tab:run_time_peak_memory}
\newcommand{\tabincell}[2]{\begin{tabular}{@{}#1@{}}#2\end{tabular}}
 \begin{center}
 \begin{threeparttable}
 \LARGE
    \resizebox{1.0\linewidth}{!}{
  \begin{tabular}{lc|cccccc}
  ~& BP & $BS\small{=}1$ & $BS\small{=}4$ &  $BS\small{=}8$ & $BS\small{=}16$ &  $BS\small{=}32$ &  $BS\small{=}64$ \\
  \cmidrule{1-8}
      NoAdapt &\ding{55} & 346  & 369  & 398  & 458  & 579  & 819  \\
      TENT & \ding{52} & 426  & 648  & 948  & 1,550  & 2,756  & 5,165  \\
      CoTTA & \ding{52} & 1,792  & 2,312  & 3,282  & 5,226  & 9,105  & 16,836 \\ 
      \methodname & \ding{55} & -- & 372  & 402  & 464  & 587  & 832  \\
      \methodname (8-bit) & \ding{55} & -- & 93  & 100  & 116  & 147  & 208  \\
\cmidrule{1-8}
~ & ~ &\multicolumn{6}{c}{$BS\small{=}1$, but update prompt every $I$ samples} \\
       ~ & ~ & $I\small{=}1$  & $I\small{=}4$  & $I\small{=}8$  & $I\small{=}16$  & $I\small{=}32$ & $I\small{=}64$  \\
\cmidrule{1-8}
      \methodname-I V1 & \ding{55} & -- &  352  & 356  & 373  & 406  & 473  \\
      \methodname-I V1  (8-bit) & \ding{55} & -- & 88  & 89  & 93  & 102  & 118 \\
      \methodname-I V2 & \ding{55} & -- & 351  & 353  & 358  & 368  & 388  \\ 
      \methodname-I V2  (8-bit)& \ding{55} & -- &  88  & 88  & 89  & 92  & 97  \\ 
 
	\end{tabular}
	}
	 \end{threeparttable}
	 \end{center}
  \vspace{-0.1in}
\end{table}

\textbf{Computational Complexity Analyses.}  
The primary computational demand of \methodname stems from $K$ (the population size in CMA) forward passes. Though \methodname requires more forward passes than TENT, its independence from backward passes significantly reduces memory usage and potentially boosts overall efficiency. In Table~\ref{tab:computation_complexity}, our \textit{Activation Shifting} achieves almost the same efficiency as NoAdapt while outperforming T3A, \ie, 59.1\% \vs~56.9\% accuracy. Notably, our BP-free \methodname ($K\small{=}2$) matches the accuracy of BP-based TENT with lower run time and memory, and \methodname ($K\small{=}6$) further surpasses BP-based CoTTA in all aspects. Moreover, even at $K\small{=}28$, \methodname maintains much higher efficiency than augmentation-based methods like MEMO~\cite{zhang2021memo}. In \methodname, $K$ is a hyperparameter that one can select different values for performance and efficiency trade-off. Note that TENT updates only norm layers, making it more efficient than full-model backpropagation.

\begin{table}[t]
    \caption{Comparisons \wrt computation complexity. \textbf{FP/BP is short forward/backward propagation.} \#FP and \#BP are numbers counted for processing a single sample. Accuracy (\%) and ECE (\%) are average results on ImageNet-C (level 5) with ViT-Base. The Wall-Clock Time (seconds) and Memory Usage (MB) are measured for processing 50,000 images of ImageNet-C on a single RTX 3090 GPU. $K$ is the population size in CMA and it works well with all $K\in[2,28]$ and $K\in\mathbb{N}^+$, as shown in Figure~\ref{fig:sensitivity} (a).}
    \label{tab:computation_complexity}
\newcommand{\tabincell}[2]{\begin{tabular}{@{}#1@{}}#2\end{tabular}}
 \begin{center}
 \begin{threeparttable}
 \LARGE
    \resizebox{1.0\linewidth}{!}{
  \begin{tabular}{l|ccc|cc|cc}
  \multicolumn{4}{c}{} & \multicolumn{2}{c}{Average} & \multicolumn{1}{c}{Run Time} & \multicolumn{1}{c}{Memory}\\
  Method & BP & \#FP & \#BP & Acc. & ECE  & (seconds) & (MB)\\
  \cmidrule{1-8}
      NoAdapt & \ding{55} &  1 & 0 & 55.5 & 10.5 & 119 & 819  \\
      T3A & \ding{55} & 1 & 0 & 56.9 & 26.8 & 235 & 957 \\
      MEMO & \ding{52} & 65 & 64 & 57.2 & 9.9 & 40,428 & 11,058 \\
      TENT & \ding{52} &  1 & 1 & 59.6 & 18.5 & 259 & 5,165\\
      SAR & \ding{52} & [1, 2] & [0, 2] & 62.7 & 7.0 & 517 & 5,166\\
      CoTTA & \ding{52} & 3or35  & 1  & 61.7 & 6.5 & 964 & 16,836\\
    \cmidrule{1-8}
      \textit{Act. Shifting} & \ding{55} & 1 & 0 & 59.1 & 12.7 & 120 & 821\\
      \methodname ($K\small{=}2$) & \ding{55} & 2 & 0 & 59.6 & 9.7 & 255 & 830\\
      \methodname ($K\small{=}4$) & \ding{55} & 4 & 0 & 60.9 & 5.8 & 497 & 830\\
      \methodname ($K\small{=}6$) & \ding{55} & 6 & 0 & 62.7 & 4.6 & 740 & 830\\
      \methodname ($K\small{=}28$) & \ding{55} & 28 & 0 & 66.3 & 3.2 & 3,386 & 832\\
	\end{tabular}
	}
	 \end{threeparttable}
	 \end{center}
  \vspace{-0.1in}
\end{table}

\textbf{Effects of Design Choice \wrt Learnable Parameters, Optimizer and Loss.} 
From Table~\ref{tab:design_choices}, directly replacing SGD with CMA for entropy-based TTA is infeasible, \eg, the average accuracy degrades from 55.5\% to 0.1\% (norm layers, exp5) and 44.9\% (prompts, exp6). The reasons are that 1) CMA fails to handle ultra-high-dimensional optimization and 2) the entropy can not provide stable learning signals and thus tends to result in collapsed trivial solutions. However, with our devised fitness function (Eqn.~(\ref{eq:fitness_function})) and learnable prompts, CMA performs effectively, surpassing the gradient-based TENT. Moreover, our proposed loss function achieves excellent performance in the context of SGD learning, \eg, a comparison between exp2 and TENT shows that the average accuracy improves significantly from 59.6\% to 70.5\%, underscoring the effectiveness of Eqn.~(\ref{eq:fitness_function}).

\begin{table}[t]
    \caption{Empirical studies of design choices \wrt learnable parameters, optimizer and loss function. We report the average results over 15 corruptions on ImageNet-C (level 5) with ViT-Base.}
    \label{tab:design_choices}
\newcommand{\tabincell}[2]{\begin{tabular}{@{}#1@{}}#2\end{tabular}}
 \begin{center}
 \begin{threeparttable}
 \LARGE
    \resizebox{1.0\linewidth}{!}{
  \begin{tabular}{lccc|cc}
        ~ & Learnable Params & Optimizer & Loss & Acc. ($\uparrow$) & ECE ($\downarrow$) \\ 
        \cmidrule{1-6}
        NoAdapt & -- & -- & -- & 55.5 & 10.5 \\ 
        TENT & norm layers & SGD & entropy & 59.6 & 18.5 \\ 
        exp1 & prompts & SGD & entropy & 50.7 & 18.4 \\ 
        exp2 & norm layers & SGD & Eqn. (\ref{eq:fitness_function}) & 70.5 & 7.9 \\ 
        exp3 & prompts & SGD & Eqn. (\ref{eq:fitness_function}) & 64.6 & 3.7 \\ 
        exp4 & norm layers & CMA & Eqn. (\ref{eq:fitness_function}) & 0.1 & 5.8 \\ 
        exp5 & norm layers & CMA & entropy & 0.1 & 99.0 \\ 
        exp6 & prompts & CMA & entropy & 44.9 & 36.8 \\ 
        \cmidrule{1-6}
        Ours  & prompts & CMA & Eqn. (\ref{eq:fitness_function}) & 65.4 & 3.3 \\ 
	\end{tabular}
	}
	 \end{threeparttable}
	 \end{center}
  \vspace{-0.1in}
\end{table}

\textbf{Effectiveness on ResNet~\cite{he2016deep} and VisionMamba~\cite{zhu2024vision}.}
For ResNet-50, we feed the original image to a learnable 7$\times$7 Conv layer to generate prompts with the same size as the image and then add prompts to the image as the model's input. For VisionMamba, we concatenate learnable input prompts with the patch embeddings. We optimize these learnable parameters/prompts by \methodname. From Table~\ref{tab:on_resnet_visionMamba}, \methodname achieves comparable performance with gradient-based TENT and SAR on VisionMamba. While on ResNet-50, \methodname outperforms BN Adapt~\cite{schneider2020improving} but still suffers a large performance gap compared to TENT. This is because convolution is a local operation, making it hard to add location-invariant input prompts in CNN to affect the whole network, whereas in transformers this can be achieved using concatenation to add prompts. How to enhance the forward-only adaptation performance on ConvNets is still a promising and challenging direction, and we leave this to our future work.

\begin{table}[t]
    \caption{Effectiveness of \methodname on ResNet and VisionMamba~\cite{zhu2024vision}. Results obtained on ImageNet-C (Gaussian noise, level 5). FOA$^\dagger$ is modified from \methodname by replacing CMA optimizer with SGD and updating the affine parameters of norm layers.}
    \label{tab:on_resnet_visionMamba}
\newcommand{\tabincell}[2]{\begin{tabular}{@{}#1@{}}#2\end{tabular}}
 \begin{center}
 \begin{threeparttable}
 \LARGE
    \resizebox{1.0\linewidth}{!}{
  \begin{tabular}{lc|cc|cc}
  \multicolumn{2}{c}{} & \multicolumn{2}{c}{ResNet-50} & \multicolumn{2}{c}{VisionMamba} \\
  Method & Need BP? & Acc. (\%) & ECE (\%) & Acc. (\%) & ECE (\%)  \\
    \cmidrule{1-6}
    NoAdapt	  & \ding{55} & 3.0	 & 19.7	& 40.9	& 3.8 \\
    BN Adapt & \ding{55} & 16.0 &	1.3	& n/a	& n/a \\
    TENT      &	\ding{52} & 29.4 & 11.4 & 49.2	& 12.1 \\
    SAR	&\ding{52} &30.7 &3.4 &49.0 & 11.4 \\
    \methodname (ours)& \ding{55} & 22.6 & 1.7 & 49.6 & 4.3 \\
    FOA$^\dagger$ & \ding{52} & 33.6 & 12.8 & 56.5 &13.6 \\
	\end{tabular}
	}
	 \end{threeparttable}
	 \end{center}
  \vspace{-0.1in}
\end{table}

\textbf{Effectiveness under Non-i.i.d. Scenarios.}
We verify the effectiveness of \methodname under two non-i.i.d. scenarios by following NOTE~\cite{gong2022note} and SAR~\cite{niu2023towards}, \ie,
\textit{online imbalanced label distribution shifts}: test data come in a class order and \textit{mixed domain shifts}: test data stream consists of multiple randomly mixed domains with different distribution shifts.
Results in Table~\ref{tab:noniid_results} show that \methodname performs stably even under non-i.i.d. settings. In general, the performance of all methods degrades when encountering non-i.i.d. scenarios. However, \methodname still achieves the best performance compared with TENT and SAR, further demonstrating the superiority of the proposed back-to-source feature alignment fitness and feature shifting scheme.

\begin{table}[t]
    \caption{Effectiveness of \methodname under non-i.i.d. scenarios. Results obtained on ViT and ImageNet-C (level 5). For mild (i.i.d.) and online imbalanced label shift scenarios, we report the average result over 15 corruptions. For mixed shifts, the performance is evaluated on a single data stream consisting of 15 mixed corruptions.}
    \label{tab:noniid_results}
\newcommand{\tabincell}[2]{\begin{tabular}{@{}#1@{}}#2\end{tabular}}
 \begin{center}
 \begin{threeparttable}
 \LARGE
    \resizebox{1.0\linewidth}{!}{
  \begin{tabular}{lc|cc|cc|cc}
  \multicolumn{2}{c}{} & \multicolumn{2}{c}{mild scenarios} & \multicolumn{2}{c}{online label shifts} & \multicolumn{2}{c}{mixed shifts} \\
  Method & BP? & Acc. & ECE & Acc. & ECE & Acc. & ECE  \\
    \cmidrule{1-8}
    TENT & \ding{52} & 59.6 & 18.5 & 60.2 & 17.7 & 56.9 & 29.2 \\
    SAR & \ding{52} & 62.7 & 7.0 & 60.8 & 7.5 & 61.4 & 14.8 \\
    \methodname (ours) & \ding{55} & \textbf{66.3} & \textbf{3.2} & \textbf{62.1} & \textbf{6.6} & \textbf{62.0} & \textbf{4.9} \\
	\end{tabular}
	}
	 \end{threeparttable}
	 \end{center}
  \vspace{-0.15in}
\end{table}

\textbf{Comparison \wrt In-Distribution Performance.}
From Table~\ref{tab:id_performance}, our \methodname maintains almost the same in-distribution accuracy as NoAdapt, outperforming all other compared methods. This success mainly benefits from: 1) we do not modify any internal model parameters, and 2) we regularize the target test features back to the source distribution via an alignment-based fitness function and an activation shifting scheme. Both of these two components are able to alleviate the issue of catastrophic forgetting.

\begin{table}[t]
    \caption{Comparison \wrt in-distribution performance, \ie, on clean/original ImageNet validation set, with ViT as the base model.}
    \label{tab:id_performance}
\newcommand{\tabincell}[2]{\begin{tabular}{@{}#1@{}}#2\end{tabular}}
 \begin{center}
 \begin{threeparttable}
 \LARGE
    \resizebox{1.0\linewidth}{!}{
  \begin{tabular}{l|c|cccc}
  Method & NoAdapt & TENT & CoTTA & SAR & \methodname (ours)  \\
    \cmidrule{1-6}
    Acc. (\%) & 85.17 & 84.80$_{(-0.37)}$ & 83.91$_{(-1.26)}$ & 84.52$_{(-0.65)}$ & \textbf{85.11$_{(-0.06)}$} \\
    ECE (\%)  & 9.6 & 3.9 &	1.8& \textbf{0.9} & 2.1 \\
	\end{tabular}
	}
	 \end{threeparttable}
	 \end{center}
  \vspace{-0.1in}
\end{table}

\textbf{Differences from Previous Forward-Only TTA.} 
The main difference is that we explicitly exploit model feedback to facilitate optimization with learnable parameters, while prior forward-only TTA methods do not. For example, BN statistics calibration-based methods, including SITA~\cite{khurana2021sita} exploit test data to calculate the mean and variances in batch norm layers, T3A~\cite{iwasawa2021test} adjusts class prototypes for adaptation, DDA~\cite{gao2023back} conducts diffusion on input images to remove the potential distribution shifts, \etc~However, these methods do not actively learn from the model's feedback \wrt a given test sample, and thus their performance on OOD data may be limited.
In contrast, our \methodname conducts explicit input prompt learning according to the model's feedback, \ie, using the proposed fitness function consisting of prediction entropy and feature discrepancy, thereby achieving much better adaptation performance.

\section{Conclusion}

In this paper, we aim to implement online test-time adaptation without the need for backpropagation and altering the model parameters. This advancement highly broadens the scope of TTA's real-world applications, particularly in resource-limited scenarios such as smartphones and FPGAs where backpropagation is often not feasible. To this end, we propose a test-time Forward-Optimization Adaptation (\methodname) method. In \methodname, we online learn an input prompt through a covariance matrix adaptation technique, paired with a designed unsupervised fitness function to provide stable learning signals. Moreover, we devise a ``back-to-source" activation shifting scheme to directly alter the activations from the out-of-distribution domain to the source in-distribution domain, further boosting the adaptation performance. Extensive experiments on four large-scale out-of-distribution benchmarks with full precision (32-bit) and quantized (8-bit/6-bit) models verify our superiority.

\section*{Acknowledgment}
This research was supported, in part, by the Joint NTU-WeBank Research Centre on Fintech, Nanyang Technological University, Singapore, and the National Research Foundation, Singapore under its Industry Alignment Fund – Pre-positioning (IAF-PP) Funding Initiative. Any opinions, findings and conclusions or recommendations expressed in this material are those of the authors and do not reflect the views of National Research Foundation, Singapore. The computational work was partially performed on resources of the National Supercomputing Centre, Singapore.

\section*{Impact Statement}
This paper presents work whose goal is to advance the field of test-time adaptation for out-of-distribution generalization. 

The societal impact of our work lies primarily in its potential to broaden the applicability of machine learning models in real-world scenarios, and facilitate their deployment on mobile computing, edge devices, and embedded systems, where power and processing capabilities are restricted. By making advanced machine learning more accessible to devices with lower specifications, our method can help democratize the benefits of AI technology. 

Ethically, our work promotes more sustainable machine learning practices by reducing the computational overhead for model adaptation. This can contribute to lower energy consumption in machine learning deployments, aligning with environmental sustainability goals. Moreover, our method enables local model adaptation without the need to upload data to cloud servers, inherently enhancing data privacy and security.

\bibliography{main}

\begin{thebibliography}{78}
\providecommand{\natexlab}[1]{#1}
\providecommand{\url}[1]{\texttt{#1}}
\expandafter\ifx\csname urlstyle\endcsname\relax
  \providecommand{\doi}[1]{doi: #1}\else
  \providecommand{\doi}{doi: \begingroup \urlstyle{rm}\Url}\fi

\bibitem[Abuduweili et~al.(2021)Abuduweili, Li, Shi, Xu, and Dou]{abuduweili2021adaptive}
Abuduweili, A., Li, X., Shi, H., Xu, C.-Z., and Dou, D.
\newblock Adaptive consistency regularization for semi-supervised transfer learning.
\newblock In \emph{IEEE Conference on Computer Vision and Pattern Recognition}, pp.\  6923--6932, 2021.

\bibitem[Bahng et~al.(2022)Bahng, Jahanian, Sankaranarayanan, and Isola]{bahng2022exploring}
Bahng, H., Jahanian, A., Sankaranarayanan, S., and Isola, P.
\newblock Exploring visual prompts for adapting large-scale models.
\newblock \emph{arXiv preprint arXiv:2203.17274}, 2022.

\bibitem[Bartler et~al.(2022)Bartler, B{\"u}hler, Wiewel, D{\"o}bler, and Yang]{bartler2022mt3}
Bartler, A., B{\"u}hler, A., Wiewel, F., D{\"o}bler, M., and Yang, B.
\newblock Mt3: Meta test-time training for self-supervised test-time adaption.
\newblock In \emph{International Conference on Artificial Intelligence and Statistics}, pp.\  3080--3090. PMLR, 2022.

\bibitem[Berger et~al.(2021)Berger, Paschali, Glocker, and Kamnitsas]{berger2021confidence}
Berger, C., Paschali, M., Glocker, B., and Kamnitsas, K.
\newblock Confidence-based out-of-distribution detection: A comparative study and analysis.
\newblock In \emph{Uncertainty for Safe Utilization of Machine Learning in Medical Imaging, and Perinatal Imaging, Placental and Preterm Image Analysis}, pp.\  122--132. Springer, 2021.

\bibitem[Beyer \& Sendhoff(2017)Beyer and Sendhoff]{beyer2017simplify}
Beyer, H.-G. and Sendhoff, B.
\newblock Simplify your covariance matrix adaptation evolution strategy.
\newblock \emph{IEEE Transactions on Evolutionary Computation}, 21\penalty0 (5):\penalty0 746--759, 2017.

\bibitem[Boudiaf et~al.(2022)Boudiaf, Mueller, Ben~Ayed, and Bertinetto]{boudiaf2022parameter}
Boudiaf, M., Mueller, R., Ben~Ayed, I., and Bertinetto, L.
\newblock Parameter-free online test-time adaptation.
\newblock In \emph{IEEE Conference on Computer Vision and Pattern Recognition}, pp.\  8344--8353, 2022.

\bibitem[Dass et~al.(2023)Dass, Wu, Shi, Li, Ye, Wang, and Lin]{dass2023vitality}
Dass, J., Wu, S., Shi, H., Li, C., Ye, Z., Wang, Z., and Lin, Y.
\newblock Vitality: Unifying low-rank and sparse approximation for vision transformer acceleration with a linear taylor attention.
\newblock In \emph{IEEE International Symposium on High-Performance Computer Architecture}, pp.\  415--428, 2023.

\bibitem[Deng et~al.(2009)Deng, Dong, Socher, Li, Li, and Fei-Fei]{deng2009imagenet}
Deng, J., Dong, W., Socher, R., Li, L.-J., Li, K., and Fei-Fei, L.
\newblock Imagenet: A large-scale hierarchical image database.
\newblock In \emph{IEEE Conference on Computer Vision and Pattern Recognition}, pp.\  248--255, 2009.

\bibitem[Deng et~al.(2024)Deng, Chen, Niu, Li, Zhuang, and Tan]{deng2024efficient}
Deng, Z., Chen, Z., Niu, S., Li, T., Zhuang, B., and Tan, M.
\newblock Efficient test-time adaptation for super-resolution with second-order degradation and reconstruction.
\newblock In \emph{Advances in Neural Information Processing Systems}, 2024.

\bibitem[Dosovitskiy et~al.(2021)Dosovitskiy, Beyer, Kolesnikov, Weissenborn, Zhai, Unterthiner, Dehghani, Minderer, Heigold, Gelly, Uszkoreit, and Houlsby]{dosovitskiy2021an}
Dosovitskiy, A., Beyer, L., Kolesnikov, A., Weissenborn, D., Zhai, X., Unterthiner, T., Dehghani, M., Minderer, M., Heigold, G., Gelly, S., Uszkoreit, J., and Houlsby, N.
\newblock An image is worth 16x16 words: Transformers for image recognition at scale.
\newblock In \emph{International Conference on Learning Representations}, 2021.

\bibitem[Dou et~al.(2019)Dou, Coelho~de Castro, Kamnitsas, and Glocker]{dou2019domain}
Dou, Q., Coelho~de Castro, D., Kamnitsas, K., and Glocker, B.
\newblock Domain generalization via model-agnostic learning of semantic features.
\newblock In \emph{Advances in Neural Information Processing Systems}, pp.\  6447--6458, 2019.

\bibitem[Duchi et~al.(2015)Duchi, Jordan, Wainwright, and Wibisono]{duchi2015optimal}
Duchi, J.~C., Jordan, M.~I., Wainwright, M.~J., and Wibisono, A.
\newblock Optimal rates for zero-order convex optimization: The power of two function evaluations.
\newblock \emph{IEEE Transactions on Information Theory}, 61\penalty0 (5):\penalty0 2788--2806, 2015.

\bibitem[Eastwood et~al.(2022)Eastwood, Mason, Williams, and Sch{\"o}lkopf]{eastwood2022source}
Eastwood, C., Mason, I., Williams, C., and Sch{\"o}lkopf, B.
\newblock Source-free adaptation to measurement shift via bottom-up feature restoration.
\newblock In \emph{International Conference on Learning Representations}, 2022.

\bibitem[Fleuret et~al.(2021)]{fleuret2021test}
Fleuret, F. et~al.
\newblock Test time adaptation through perturbation robustness.
\newblock In \emph{Advances in Neural Information Processing Systems Workshop}, 2021.

\bibitem[Gandelsman et~al.(2022)Gandelsman, Sun, Chen, and Efros]{gandelsman2022test}
Gandelsman, Y., Sun, Y., Chen, X., and Efros, A.
\newblock Test-time training with masked autoencoders.
\newblock In \emph{Advances in Neural Information Processing Systems}, volume~35, pp.\  29374--29385, 2022.

\bibitem[Gao et~al.(2023)Gao, Zhang, Liu, Shelhamer, Darrell, and Wang]{gao2023back}
Gao, J., Zhang, J., Liu, X., Shelhamer, E., Darrell, T., and Wang, D.
\newblock Back to the source: Diffusion-driven test-time adaptation.
\newblock In \emph{IEEE Conference on Computer Vision and Pattern Recognition}, 2023.

\bibitem[Gidaris et~al.(2018)Gidaris, Singh, and Komodakis]{gidaris2018unsupervised}
Gidaris, S., Singh, P., and Komodakis, N.
\newblock Unsupervised representation learning by predicting image rotations.
\newblock In \emph{International Conference on Learning Representations}, 2018.

\bibitem[Gong et~al.(2022)Gong, Jeong, Kim, Kim, Shin, and Lee]{gong2022note}
Gong, T., Jeong, J., Kim, T., Kim, Y., Shin, J., and Lee, S.-J.
\newblock Note: Robust continual test-time adaptation against temporal correlation.
\newblock In \emph{Advances in Neural Information Processing Systems}, volume~35, pp.\  27253--27266, 2022.

\bibitem[Goyal et~al.(2022)Goyal, Sun, Raghunathan, and Kolter]{goyal2022test}
Goyal, S., Sun, M., Raghunathan, A., and Kolter, J.~Z.
\newblock Test time adaptation via conjugate pseudo-labels.
\newblock In \emph{Advances in Neural Information Processing Systems}, volume~35, pp.\  6204--6218, 2022.

\bibitem[Hansen(2016)]{hansen2016cma}
Hansen, N.
\newblock The cma evolution strategy: A tutorial.
\newblock \emph{arXiv preprint arXiv:1604.00772}, 2016.

\bibitem[Hansen \& Ostermeier(2001)Hansen and Ostermeier]{hansen2001completely}
Hansen, N. and Ostermeier, A.
\newblock Completely derandomized self-adaptation in evolution strategies.
\newblock \emph{Evolutionary Computation}, 9\penalty0 (2):\penalty0 159--195, 2001.

\bibitem[Hansen et~al.(2003)Hansen, M{\"u}ller, and Koumoutsakos]{hansen2003reducing}
Hansen, N., M{\"u}ller, S.~D., and Koumoutsakos, P.
\newblock Reducing the time complexity of the derandomized evolution strategy with covariance matrix adaptation (cma-es).
\newblock \emph{Evolutionary Computation}, 11\penalty0 (1):\penalty0 1--18, 2003.

\bibitem[He et~al.(2016)He, Zhang, Ren, and Sun]{he2016deep}
He, K., Zhang, X., Ren, S., and Sun, J.
\newblock Deep residual learning for image recognition.
\newblock In \emph{IEEE Conference on Computer Vision and Pattern Recognition}, pp.\  770--778, 2016.

\bibitem[He et~al.(2020)He, Zheng, and Zhou]{he2020mmes}
He, X., Zheng, Z., and Zhou, Y.
\newblock Mmes: Mixture model-based evolution strategy for large-scale optimization.
\newblock \emph{IEEE Transactions on Evolutionary Computation}, 25\penalty0 (2):\penalty0 320--333, 2020.

\bibitem[Hendrycks \& Dietterich(2019)Hendrycks and Dietterich]{hendrycks2019benchmarking}
Hendrycks, D. and Dietterich, T.
\newblock Benchmarking neural network robustness to common corruptions and perturbations.
\newblock In \emph{International Conference on Learning Representations}, 2019.

\bibitem[Hendrycks et~al.(2020)Hendrycks, Mu, Cubuk, Zoph, Gilmer, and Lakshminarayanan]{hendrycks2020augmix}
Hendrycks, D., Mu, N., Cubuk, E.~D., Zoph, B., Gilmer, J., and Lakshminarayanan, B.
\newblock Augmix: {A} simple data processing method to improve robustness and uncertainty.
\newblock In \emph{International Conference on Learning Representations}, 2020.

\bibitem[Hendrycks et~al.(2021)Hendrycks, Basart, Mu, Kadavath, Wang, Dorundo, Desai, Zhu, Parajuli, Guo, et~al.]{hendrycks2021many}
Hendrycks, D., Basart, S., Mu, N., Kadavath, S., Wang, F., Dorundo, E., Desai, R., Zhu, T., Parajuli, S., Guo, M., et~al.
\newblock The many faces of robustness: A critical analysis of out-of-distribution generalization.
\newblock In \emph{IEEE Conference on Computer Vision and Pattern Recognition}, pp.\  8340--8349, 2021.

\bibitem[Hong et~al.(2023)Hong, Lyu, Zhou, and Spranger]{hong2023mecta}
Hong, J., Lyu, L., Zhou, J., and Spranger, M.
\newblock {MECTA}: Memory-economic continual test-time model adaptation.
\newblock In \emph{International Conference on Learning Representations}, 2023.

\bibitem[Hu et~al.(2021)Hu, Uzunbas, Chen, Wang, Shah, Nevatia, and Lim]{hu2021mixnorm}
Hu, X., Uzunbas, G., Chen, S., Wang, R., Shah, A., Nevatia, R., and Lim, S.-N.
\newblock Mixnorm: Test-time adaptation through online normalization estimation.
\newblock \emph{arXiv preprint arXiv:2110.11478}, 2021.

\bibitem[Iwasawa \& Matsuo(2021)Iwasawa and Matsuo]{iwasawa2021test}
Iwasawa, Y. and Matsuo, Y.
\newblock Test-time classifier adjustment module for model-agnostic domain generalization.
\newblock In \emph{Advances in Neural Information Processing Systems}, volume~34, 2021.

\bibitem[Jamieson et~al.(2012)Jamieson, Nowak, and Recht]{jamieson2012query}
Jamieson, K.~G., Nowak, R., and Recht, B.
\newblock Query complexity of derivative-free optimization.
\newblock In \emph{Advances in Neural Information Processing Systems}, volume~25, 2012.

\bibitem[Jia et~al.(2022)Jia, Tang, Chen, Cardie, Belongie, Hariharan, and Lim]{jia2022visual}
Jia, M., Tang, L., Chen, B.-C., Cardie, C., Belongie, S., Hariharan, B., and Lim, S.-N.
\newblock Visual prompt tuning.
\newblock In \emph{European Conference on Computer Vision}, pp.\  709--727. Springer, 2022.

\bibitem[Khurana et~al.(2021)Khurana, Paul, Rai, Biswas, and Aggarwal]{khurana2021sita}
Khurana, A., Paul, S., Rai, P., Biswas, S., and Aggarwal, G.
\newblock Sita: Single image test-time adaptation.
\newblock \emph{arXiv preprint arXiv:2112.02355}, 2021.

\bibitem[Koh et~al.(2021)Koh, Sagawa, Marklund, Xie, Zhang, Balsubramani, Hu, Yasunaga, Phillips, Gao, et~al.]{koh2021wilds}
Koh, P.~W., Sagawa, S., Marklund, H., Xie, S.~M., Zhang, M., Balsubramani, A., Hu, W., Yasunaga, M., Phillips, R.~L., Gao, I., et~al.
\newblock Wilds: A benchmark of in-the-wild distribution shifts.
\newblock In \emph{International Conference on Machine Learning}, pp.\  5637--5664, 2021.

\bibitem[Li et~al.(2023{\natexlab{a}})Li, Hopkins, Bau, Vi{\'e}gas, Pfister, and Wattenberg]{li2022emergent}
Li, K., Hopkins, A.~K., Bau, D., Vi{\'e}gas, F., Pfister, H., and Wattenberg, M.
\newblock Emergent world representations: Exploring a sequence model trained on a synthetic task.
\newblock In \emph{International Conference on Learning Representations}, 2023{\natexlab{a}}.

\bibitem[Li et~al.(2023{\natexlab{b}})Li, Patel, Viégas, Pfister, and Wattenberg]{li2023inferencetime}
Li, K., Patel, O., Viégas, F., Pfister, H., and Wattenberg, M.
\newblock Inference-time intervention: Eliciting truthful answers from a language model.
\newblock In \emph{Advances in Neural Information Processing Systems}, 2023{\natexlab{b}}.

\bibitem[Liang et~al.(2023)Liang, He, and Tan]{liang2023comprehensive}
Liang, J., He, R., and Tan, T.
\newblock A comprehensive survey on test-time adaptation under distribution shifts.
\newblock \emph{arXiv preprint arXiv:2303.15361}, 2023.

\bibitem[Lim et~al.(2023)Lim, Kim, Choo, and Choi]{lim2023ttn}
Lim, H., Kim, B., Choo, J., and Choi, S.
\newblock {TTN}: A domain-shift aware batch normalization in test-time adaptation.
\newblock In \emph{International Conference on Learning Representations}, 2023.

\bibitem[Lin et~al.(2023)Lin, Mirza, Kozinski, Possegger, Kuehne, and Bischof]{lin2023video}
Lin, W., Mirza, M.~J., Kozinski, M., Possegger, H., Kuehne, H., and Bischof, H.
\newblock Video test-time adaptation for action recognition.
\newblock In \emph{IEEE Conference on Computer Vision and Pattern Recognition}, pp.\  22952--22961, 2023.

\bibitem[Liu et~al.(2021{\natexlab{a}})Liu, Kothari, van Delft, Bellot-Gurlet, Mordan, and Alahi]{liu2021ttt++}
Liu, Y., Kothari, P., van Delft, B., Bellot-Gurlet, B., Mordan, T., and Alahi, A.
\newblock Ttt++: When does self-supervised test-time training fail or thrive?
\newblock In \emph{Advances in Neural Information Processing Systems}, volume~34, 2021{\natexlab{a}}.

\bibitem[Liu et~al.(2021{\natexlab{b}})Liu, Wang, Han, Zhang, Ma, and Gao]{liu2021post}
Liu, Z., Wang, Y., Han, K., Zhang, W., Ma, S., and Gao, W.
\newblock Post-training quantization for vision transformer.
\newblock In \emph{Advances in Neural Information Processing Systems}, volume~34, pp.\  28092--28103, 2021{\natexlab{b}}.

\bibitem[Long et~al.(2015)Long, Cao, Wang, and Jordan]{long2015learning}
Long, M., Cao, Y., Wang, J., and Jordan, M.
\newblock Learning transferable features with deep adaptation networks.
\newblock In \emph{International Conference on Machine Learning}, pp.\  97--105. PMLR, 2015.

\bibitem[Long et~al.(2016)Long, Zhu, Wang, and Jordan]{long2016unsupervised}
Long, M., Zhu, H., Wang, J., and Jordan, M.~I.
\newblock Unsupervised domain adaptation with residual transfer networks.
\newblock In \emph{Advances in Neural Information Processing Systems}, volume~29, 2016.

\bibitem[Loshchilov et~al.(2018)Loshchilov, Glasmachers, and Beyer]{loshchilov2018large}
Loshchilov, I., Glasmachers, T., and Beyer, H.-G.
\newblock Large scale black-box optimization by limited-memory matrix adaptation.
\newblock \emph{IEEE Transactions on Evolutionary Computation}, 23\penalty0 (2):\penalty0 353--358, 2018.

\bibitem[Louizos et~al.(2019)Louizos, Reisser, Blankevoort, Gavves, and Welling]{Louizos2019relaxed}
Louizos, C., Reisser, M., Blankevoort, T., Gavves, E., and Welling, M.
\newblock Relaxed quantization for discretized neural networks.
\newblock In \emph{International Conference on Learning Representations}, 2019.

\bibitem[Malladi et~al.(2023)Malladi, Gao, Nichani, Damian, Lee, Chen, and Arora]{malladi2023fine}
Malladi, S., Gao, T., Nichani, E., Damian, A., Lee, J.~D., Chen, D., and Arora, S.
\newblock Fine-tuning language models with just forward passes.
\newblock In \emph{Advances in Neural Information Processing Systems}, 2023.

\bibitem[Mirza et~al.(2023)Mirza, Soneira, Lin, Kozinski, Possegger, and Bischof]{mirza2023actmad}
Mirza, M.~J., Soneira, P.~J., Lin, W., Kozinski, M., Possegger, H., and Bischof, H.
\newblock Actmad: Activation matching to align distributions for test-time-training.
\newblock In \emph{IEEE Conference on Computer Vision and Pattern Recognition}, pp.\  24152--24161, 2023.

\bibitem[Nado et~al.(2020)Nado, Padhy, Sculley, D'Amour, Lakshminarayanan, and Snoek]{nado2020evaluating}
Nado, Z., Padhy, S., Sculley, D., D'Amour, A., Lakshminarayanan, B., and Snoek, J.
\newblock Evaluating prediction-time batch normalization for robustness under covariate shift.
\newblock \emph{arXiv preprint arXiv:2006.10963}, 2020.

\bibitem[Naeini et~al.(2015)Naeini, Cooper, and Hauskrecht]{naeini2015ece}
Naeini, M.~P., Cooper, G., and Hauskrecht, M.
\newblock Obtaining well calibrated probabilities using bayesian binning.
\newblock In \emph{AAAI Conference on Artificial Intelligence}, volume~29, 2015.

\bibitem[Niu et~al.(2022)Niu, Wu, Zhang, Chen, Zheng, Zhao, and Tan]{niu2022efficient}
Niu, S., Wu, J., Zhang, Y., Chen, Y., Zheng, S., Zhao, P., and Tan, M.
\newblock Efficient test-time model adaptation without forgetting.
\newblock In \emph{International Conference on Machine Learning}, pp.\  16888--16905, 2022.

\bibitem[Niu et~al.(2023)Niu, Wu, Zhang, Wen, Chen, Zhao, and Tan]{niu2023towards}
Niu, S., Wu, J., Zhang, Y., Wen, Z., Chen, Y., Zhao, P., and Tan, M.
\newblock Towards stable test-time adaptation in dynamic wild world.
\newblock In \emph{International Conference on Learning Representations}, 2023.

\bibitem[Oh et~al.(2024)Oh, Lee, Choi, Jung, Hwang, and Yoon]{oh2024efficient}
Oh, Y., Lee, J., Choi, J., Jung, D., Hwang, U., and Yoon, S.
\newblock Efficient diffusion-driven corruption editor for test-time adaptation.
\newblock \emph{arXiv preprint arXiv:2403.10911}, 2024.

\bibitem[Qiu et~al.(2021)Qiu, Zhang, Lin, Niu, Liu, Du, and Tan]{Qiu2021CPGA}
Qiu, Z., Zhang, Y., Lin, H., Niu, S., Liu, Y., Du, Q., and Tan, M.
\newblock Source-free domain adaptation via avatar prototype generation and adaptation.
\newblock In \emph{International Joint Conference on Artificial Intelligence}, 2021.

\bibitem[Recht et~al.(2019)Recht, Roelofs, Schmidt, and Shankar]{recht2019imagenet}
Recht, B., Roelofs, R., Schmidt, L., and Shankar, V.
\newblock Do imagenet classifiers generalize to imagenet?
\newblock In \emph{International Conference on Machine Learning}, pp.\  5389--5400, 2019.

\bibitem[Saito et~al.(2018)Saito, Watanabe, Ushiku, and Harada]{saito2018maximum}
Saito, K., Watanabe, K., Ushiku, Y., and Harada, T.
\newblock Maximum classifier discrepancy for unsupervised domain adaptation.
\newblock In \emph{IEEE Conference on Computer Vision and Pattern Recognition}, pp.\  3723--3732, 2018.

\bibitem[Schneider et~al.(2020)Schneider, Rusak, Eck, Bringmann, Brendel, and Bethge]{schneider2020improving}
Schneider, S., Rusak, E., Eck, L., Bringmann, O., Brendel, W., and Bethge, M.
\newblock Improving robustness against common corruptions by covariate shift adaptation.
\newblock In \emph{Advances in Neural Information Processing Systems}, volume~33, pp.\  11539--11551, 2020.

\bibitem[Shamir(2017)]{shamir2017optimal}
Shamir, O.
\newblock An optimal algorithm for bandit and zero-order convex optimization with two-point feedback.
\newblock \emph{The Journal of Machine Learning Research}, 18\penalty0 (1):\penalty0 1703--1713, 2017.

\bibitem[Shankar et~al.(2018)Shankar, Piratla, Chakrabarti, Chaudhuri, Jyothi, and Sarawagi]{shankar2018generalizing}
Shankar, S., Piratla, V., Chakrabarti, S., Chaudhuri, S., Jyothi, P., and Sarawagi, S.
\newblock Generalizing across domains via cross-gradient training.
\newblock In \emph{International Conference on Learning Representations}, 2018.

\bibitem[Shu et~al.(2022)Shu, Nie, Huang, Yu, Goldstein, Anandkumar, and Xiao]{shu2022test}
Shu, M., Nie, W., Huang, D.-A., Yu, Z., Goldstein, T., Anandkumar, A., and Xiao, C.
\newblock Test-time prompt tuning for zero-shot generalization in vision-language models.
\newblock In \emph{Advances in Neural Information Processing Systems}, volume~35, pp.\  14274--14289, 2022.

\bibitem[Subramani et~al.(2022)Subramani, Suresh, and Peters]{subramani2022extracting}
Subramani, N., Suresh, N., and Peters, M.~E.
\newblock Extracting latent steering vectors from pretrained language models.
\newblock In \emph{Findings of the Association for Computational Linguistics}, 2022.

\bibitem[Sun et~al.(2022)Sun, Shao, Qian, Huang, and Qiu]{sun2022black}
Sun, T., Shao, Y., Qian, H., Huang, X., and Qiu, X.
\newblock Black-box tuning for language-model-as-a-service.
\newblock In \emph{International Conference on Machine Learning}, pp.\  20841--20855. PMLR, 2022.

\bibitem[Sun et~al.(2020)Sun, Wang, Liu, Miller, Efros, and Hardt]{sun2020test}
Sun, Y., Wang, X., Liu, Z., Miller, J., Efros, A., and Hardt, M.
\newblock Test-time training with self-supervision for generalization under distribution shifts.
\newblock In \emph{International Conference on Machine Learning}, pp.\  9229--9248, 2020.

\bibitem[Tan et~al.(2024)Tan, Chen, Wu, Zhang, Chen, Zhao, and Niu]{tan2024uncertainty}
Tan, M., Chen, G., Wu, J., Zhang, Y., Chen, Y., Zhao, P., and Niu, S.
\newblock Uncertainty-calibrated test-time model adaptation without forgetting.
\newblock \emph{arXiv preprint arXiv:2403.11491}, 2024.

\bibitem[Turner et~al.(2023)Turner, Thiergart, Udell, Leech, Mini, and MacDiarmid]{turner2023activation}
Turner, A., Thiergart, L., Udell, D., Leech, G., Mini, U., and MacDiarmid, M.
\newblock Activation addition: Steering language models without optimization.
\newblock \emph{arXiv preprint arXiv:2308.10248}, 2023.

\bibitem[Wang et~al.(2021)Wang, Shelhamer, Liu, Olshausen, and Darrell]{wang2021tent}
Wang, D., Shelhamer, E., Liu, S., Olshausen, B., and Darrell, T.
\newblock Tent: Fully test-time adaptation by entropy minimization.
\newblock In \emph{International Conference on Learning Representations}, 2021.

\bibitem[Wang et~al.(2019)Wang, Ge, Lipton, and Xing]{wang2019learning}
Wang, H., Ge, S., Lipton, Z., and Xing, E.~P.
\newblock Learning robust global representations by penalizing local predictive power.
\newblock In \emph{Advances in Neural Information Processing Systems}, pp.\  10506--10518, 2019.

\bibitem[Wang et~al.(2022)Wang, Fink, Van~Gool, and Dai]{wang2022continual}
Wang, Q., Fink, O., Van~Gool, L., and Dai, D.
\newblock Continual test-time domain adaptation.
\newblock In \emph{IEEE Conference on Computer Vision and Pattern Recognition}, 2022.

\bibitem[Wen et~al.(2023)Wen, Niu, Li, Wu, Tan, and Wu]{wen2023test}
Wen, Z., Niu, S., Li, G., Wu, Q., Tan, M., and Wu, Q.
\newblock Test-time model adaptation for visual question answering with debiased self-supervisions.
\newblock \emph{IEEE Transactions on Multimedia}, 2023.

\bibitem[Wightman(2019)]{rw2019timm}
Wightman, R.
\newblock Pytorch image models.
\newblock \url{https://github.com/rwightman/pytorch-image-models}, 2019.

\bibitem[Yao et~al.(2022)Yao, Wang, Li, Zhang, Liang, Zou, and Finn]{yao2022improving}
Yao, H., Wang, Y., Li, S., Zhang, L., Liang, W., Zou, J., and Finn, C.
\newblock Improving out-of-distribution robustness via selective augmentation.
\newblock In \emph{International Conference on Machine Learning}, 2022.

\bibitem[You et~al.(2023)You, Sun, Shi, Yu, Zhao, Zhang, Li, Li, and Lin]{you2023vitcod}
You, H., Sun, Z., Shi, H., Yu, Z., Zhao, Y., Zhang, Y., Li, C., Li, B., and Lin, Y.
\newblock Vitcod: Vision transformer acceleration via dedicated algorithm and accelerator co-design.
\newblock In \emph{IEEE International Symposium on High-Performance Computer Architecture}, pp.\  273--286, 2023.

\bibitem[Yu et~al.(2023)Yu, Chen, Lin, and He]{yu2023black}
Yu, L., Chen, Q., Lin, J., and He, L.
\newblock Black-box prompt tuning for vision-language model as a service.
\newblock In \emph{International Joint Conference on Artificial Intelligence}, pp.\  1686--1694, 2023.

\bibitem[Yuan et~al.(2022)Yuan, Xue, Chen, Wu, and Sun]{yuan2022ptq4vit}
Yuan, Z., Xue, C., Chen, Y., Wu, Q., and Sun, G.
\newblock Ptq4vit: Post-training quantization for vision transformers with twin uniform quantization.
\newblock In \emph{European Conference on Computer Vision}, pp.\  191--207. Springer, 2022.

\bibitem[Zeng et~al.(2023)Zeng, Deng, Xu, Niu, and Chen]{zeng2023exploring}
Zeng, R., Deng, Q., Xu, H., Niu, S., and Chen, J.
\newblock Exploring motion cues for video test-time adaptation.
\newblock In \emph{Proceedings of the 31st ACM International Conference on Multimedia}, pp.\  1840--1850, 2023.

\bibitem[Zhang et~al.(2022)Zhang, Levine, and Finn]{zhang2021memo}
Zhang, M.~M., Levine, S., and Finn, C.
\newblock Memo: Test time robustness via adaptation and augmentation.
\newblock In \emph{Advances in Neural Information Processing Systems}, 2022.

\bibitem[Zhang et~al.(2020)Zhang, Wei, Wu, Zhao, Niu, Huang, and Tan]{zhang2020collaborative}
Zhang, Y., Wei, Y., Wu, Q., Zhao, P., Niu, S., Huang, J., and Tan, M.
\newblock Collaborative unsupervised domain adaptation for medical image diagnosis.
\newblock \emph{IEEE Transactions on Image Processing}, 29:\penalty0 7834--7844, 2020.

\bibitem[Zhao et~al.(2023)Zhao, Chen, and Xia]{zhao2023delta}
Zhao, B., Chen, C., and Xia, S.-T.
\newblock {DELTA}: Degradation-free fully test-time adaptation.
\newblock In \emph{International Conference on Learning Representations}, 2023.

\bibitem[Zhu et~al.(2024)Zhu, Liao, Zhang, Wang, Liu, and Wang]{zhu2024vision}
Zhu, L., Liao, B., Zhang, Q., Wang, X., Liu, W., and Wang, X.
\newblock Vision mamba: Efficient visual representation learning with bidirectional state space model.
\newblock \emph{arXiv preprint arXiv:2401.09417}, 2024.

\end{thebibliography}
\bibliographystyle{icml2024}

\newpage
\appendix

\twocolumn[\icmltitle{Test-Time Model Adaptation with Only Forward Passes \\ \texttt{Supplementary Materials}}]{}

\section{Related Work}\label{sec:related_work}

\textbf{Test-Time Adaptation} (TTA) is designed to enhance the performance of a model on unseen test data, which may exhibit distribution shifts, by directly learning from the test data itself. We categorize the related TTA works into the following two groups for discussion, differentiated by their dependence on backward propagation.

\textit{\textbf{$\bullet$ Backpropagation (BP)-Free TTA.}} In the early development of BP-free TTA, attention was primarily given to adjusting batch normalization (BN) layer statistics by computing mean and variance from testing data~\cite{nado2020evaluating,schneider2020improving,gong2022note}. This method, however, assumes multiple test samples for each prediction. To address this, later studies proposed single-sample BN adaptation techniques, such as using data augmentation~\cite{khurana2021sita}, mix-up training and testing statistics~\cite{hu2021mixnorm,lim2023ttn}. Moreover, NOTE~\cite{gong2022note} proposes instance-aware batch normalization and DELTA~\cite{zhao2023delta} exploits test-time batch re-normalization to adapt a given model under non-i.i.d. testing scenarios. In addition to BN adaptation, other methodologies have been explored and shown to be effective, such as prototype-based classifier adjustment~\cite{iwasawa2021test}, predicted logits correction~\cite{boudiaf2022parameter}, diffusion-based input image adaptation~\cite{gao2023back,oh2024efficient}. However, since BP-free TTA does not update the core model parameters, it might exhibit limited learning capabilities, often resulting in suboptimal performance when dealing with out-of-distribution testing data. Therefore, BP-based TTA emerged as an effective solution to significantly boost the out-of-distribution generalization performance.

\textit{\textbf{$\bullet$ Backpropagation-Based TTA.}} One pioneering work of BP-based TTA is known as Test-Time Training (TTT)~\cite{sun2020test}. For TTT methods, they first train a source model using both supervised and self-supervised objectives, followed by adapting the model at test time with the self-supervised objective, such as rotation prediction~\cite{sun2020test}, contrastive learning~\cite{liu2021ttt++,bartler2022mt3}, reconstruction learning~\cite{gandelsman2022test,deng2024efficient}. To avoid directly altering the model training process and access to source data, Fully TTA methods update any given model via unsupervised learning objectives, such as entropy minimization~\cite{wang2021tent,wen2023test,niu2023towards,tan2024uncertainty}, prediction consistency maximization~\cite{zhang2021memo,fleuret2021test,zeng2023exploring} and feature distribution alignment~\cite{mirza2023actmad,lin2023video}. 

However, BP-based TTA methods typically require multiple backward propagations for each test sample, leading to computational inefficiency. To tackle this, recent works~\cite{niu2022efficient,niu2023towards,wang2022continual,shu2022test} have proposed selecting confident or reliable samples for test-time model learning. This strategy significantly reduces the number of backward passes needed for an entire set of testing data, thereby enhancing adaptation efficiency and performance. Moreover, MECTA~\cite{hong2023mecta} proposes a series of techniques to reduce the run-time memory consumption of BP-based TTA, including reducing the batch size and stopping the BP caching heuristically. Nonetheless, these TTA methods still depend on BP, which poses challenges for resource-constrained edge devices like smartphones and FPGAs, especially those with quantized models. These devices often have limited memory and may not support backpropagation, thus hindering the broad real-world applications of BP-based TTA methods. In this context, we introduce a forward-only and optimization-based TTA method, aiming to achieve better performance than BP-based TTA methods but without using any backward propagation.

\textbf{Derivative-Free Learning} (DFL) achieves optimization solely through evaluating the function values $f(\bx)$ at sampled solutions $\bx$, with minimal assumptions about the objective function (can be non-convex or black-box) and without the need for explicit derivatives. We review representative DFL methods below.

\textit{\textbf{$\bullet$ Evolutionary Algorithms.}} Covariance Matrix Adaptation Evolution Strategy (CMA-ES)~\cite{hansen2016cma} is one of the most successful evolutionary approaches. It updates a covariance matrix of the multivariate normal distribution used to sample new solutions and effectively learns a second-order model of the objective function, similar to approximating the inverse Hessian matrix in classical optimization. Based on CMA-ES,  several recent works~\cite{beyer2017simplify,he2020mmes,loshchilov2018large} have been proposed to improve CMA-ES's time and memory complexity.

Nevertheless, CMA-ES still faces challenges in handling extremely high-dimensional optimization problems, such as deep model optimization. Recently, some efforts~\cite{sun2022black,yu2023black} have been made to adapt CMA algorithms for the scenario of Model-as-Service applications in language and vision-language models. However, these methods run on a pre-collected annotated dataset for a specific downstream task, and perform supervised optimization offline. In contrast, our approach adapts a given model to out-of-distribution testing data in an online and unsupervised manner.

\begin{figure*}[t]
    \centering
    \includegraphics[width=0.88\linewidth]{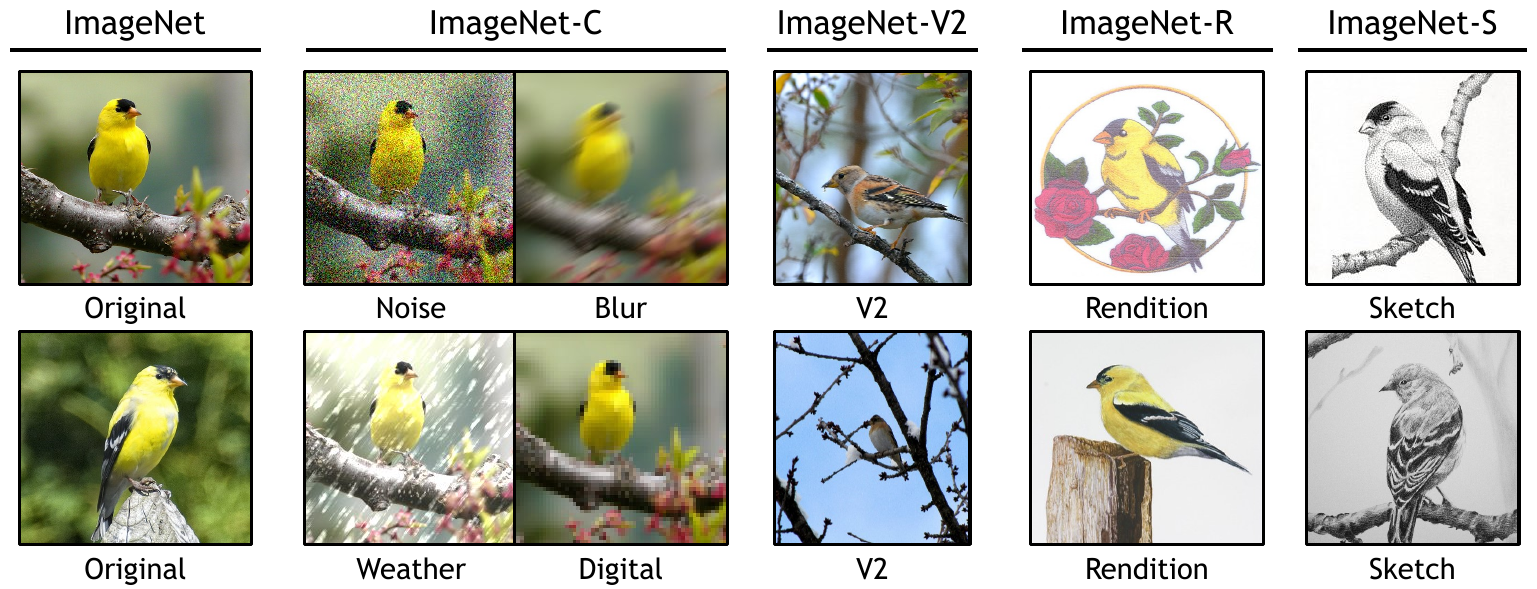}
    \vspace{-0.1in}
    \caption{Visualizations of images in ImageNet and ImageNet-C/V2/R/Sketch, which are directly taken from their original papers.}
    \label{fig:dataset_images}
\end{figure*}

\textit{\textbf{$\bullet$ Zeroth-Order Optimization}} methods estimate gradients by comparing the loss values of two forward passes~\cite{duchi2015optimal,jamieson2012query,shamir2017optimal}, in which the model weights used in the second forward pass are altered from that used in the first forward pass through a random perturbation. One of the recent works~\cite{malladi2023fine} introduced this optimization scheme in fine-tuning large language models and demonstrated its effectiveness. However, these methods still operate in an offline setting with few-shot supervised samples and are theorized to have slow convergence for optimizing large models~\cite{malladi2023fine}, which are not directly compatible with our online unsupervised TTA setting. Thus, in our \methodname, we adopt the evolution-based CMA-ES method for prompt optimization. However, it is important to note that Zeroth-Order optimization also presents a promising avenue for developing backpropagation-free TTA methods. Exploring this potential is an objective we aim to pursue in the future.

{\textbf{Activation Editing}} methods directly modify the internal activation representations of a model to control the final output~\cite{li2022emergent,subramani2022extracting}, which has recently been explored in the field of language models. For example, \citet{subramani2022extracting} and \citet{turner2023activation} exploit ``steering vectors" for style transfer and \citet{li2023inferencetime} adjust the attention heads to improve the truthfulness of large language models. In this work, inspired by this general idea, we propose a back-to-source activation shifting mechanism that online adjusts the model's activations in real-time to enhance generalization to out-of-distribution testing data.

\textit{Connection to Conventional Alignment-based Domain Adaptation Methods.} Conventional DA methods often seek to learn domain-invariant features by minimizing the Maximum Mean Discrepancy (MMD) between the source and target distributions~\cite{long2015learning,long2016unsupervised,abuduweili2021adaptive}. BUFR~\cite{eastwood2022source} extends this idea to a source-free manner and directly minimizes the KL divergence between the source and target feature distributions. Our proposed Activation Shifting scheme follows the general idea of source-target alignment but differs from them mainly in two aspects. 1) Conventional DA methods often calculate the MMD or KL divergence as a differentiable loss function and exploit it for gradient backpropagation, thereby updating model weights. In contrast, our Activation Shifting directly edits the activations back to the source distribution in a gradient-free manner without modifying model weights, which facilitates efficient test-time adaptation. 2) Previous DA methods typically work offline, \ie, training the model on a pre-collected dataset for several epochs. Conversely, our Activation Shifting approach is effective for online test data streams, where adaptation occurs immediately after each mini-batch of test samples is processed (\eg, with batch sizes of 2 or 4).

\section{More Implementation Details}
\label{suppl:sec:implentation}
\subsection{More Details on Dataset}
In this paper, we conduct experiments on four ImageNet~\cite{deng2009imagenet} variants to evaluate the out-of-distribution generalization ability, \ie, ImageNet-C~\cite{hendrycks2019benchmarking}, ImageNet-R~\cite{hendrycks2021many}, ImageNet-V2~\cite{recht2019imagenet}, and ImageNet-Sketch~\cite{wang2019learning}.

\textbf{ImageNet-C} consists of various versions of corruption applied to 50,000 validation images from ImageNet. The dataset encompasses 15 distinct corruption types of 4 main groups, including Gaussian noise, shot noise, impulse noise, defocus blur, glass blur, motion blur, zoom blur, snow, frost, fog, brightness, contrast, elastic transformation, pixelation, and JPEG compression. Each corruption type is characterized by 5 different levels of severity, with higher severity levels indicating a more severe distribution shift. In our experiments, we specifically utilize severity level 5 for evaluation.

\textbf{ImageNet-R} contains 30,000 images featuring diverse artistic renditions of 200 ImageNet classes. These images are predominantly sourced from Flickr and subjected to filtering by Amazon MTurk annotators.

\textbf{ImageNet-V2} is a newly collected test dataset extracted from the same test distribution as ImageNet. It comprises three test sets, each containing 10,000 new images and covering 1000 ImageNet classes. Following previous TTA methods~\cite{nado2020evaluating}, we utilize the Matched Frequency subset of ImageNet-V2 for evaluation, in which the images are sampled to match the class frequency distributions of the original ImageNet validation dataset.

\textbf{ImageNet-Sketch} consists of 50,899 images represented as black and white sketches, encompassing 1000 ImageNet classes. Each class contains approximately 50 images.

\subsection{More Evaluation Protocols}
\label{suppl:subsec:experiment protocols}
We use ViT-Base~\cite{dosovitskiy2021an} as the source model backbone for all experiments. The model is trained on the source ImageNet-1K training set and the model weights\footnote{\href{https://storage.googleapis.com/vit_models/augreg/B_16-i21k-300ep-lr_0.001-aug_medium1-wd_0.1-do_0.0-sd_0.0--imagenet2012-steps_20k-lr_0.01-res_224.npz}{https://storage.googleapis.com/vit\_models/augreg/B\_16-i21k-300ep-lr\_0.001-aug\_medium1-wd\_0.1-do\_0.0-sd\_0.0--imagenet2012-steps\_20k-lr\_0.01-res\_224.npz}} are directly obtained from \texttt{timm}\footnote{\href{https://github.com/pprp/timm}{https://github.com/pprp/timm}} repository~\cite{rw2019timm}. We adopt PTQ4ViT\footnote{\href{https://github.com/hahnyuan/PTQ4ViT}{https://github.com/hahnyuan/PTQ4ViT}}~\cite{yuan2022ptq4vit} for 8-bit and 6-bit model quantization with 32 randomly selected samples from the training set. We introduce the implementation details of all involved methods in the following.

\textbf{\methodname (Ours).} For the default configuration of hyperparameters, the number of prompt embeddings $N_p$ is set to 3 using uniform initialization. We use CMA-ES\footnote{\href{https://github.com/CMA-ES/pycma}{https://github.com/CMA-ES/pycma}} as the update rule, with the batch size $BS$ of 64, and the population size $K$ of $28=4+3\times\log(prompt ~~dim)$ by following~\cite{hansen2016cma}. The $\lambda$ in Eqn.~(\ref{eq:fitness_function}) is set to 0.4$\times BS/64$ on ImageNet-C/V2/Sketch, and 0.2$\times BS/64$ on ImageNet-R to balance the magnitude of two losses. We use the validation set of ImageNet-1K to estimate source training statistics. The moving average factor $\alpha$ in Eqn.~(\ref{eq:testing_mu}) is set to 0.1. The step size $\gamma$ in Eqn.~(\ref{eq:activation shifting}) is set to 1.0, aiming to exactly align the overall center of testing and training features.  For updating prompts and normalization layers with SGD optimizer in Table~\ref{tab:design_choices} with  Eqn.~(\ref{eq:fitness_function}), the entropy loss is divided by the batch size of 64 and the $\lambda$ is set to 30 to make the two losses have similar magnitude. We investigate the effects of these hyperparameters with different setups in Section~\ref{sec:ablation} and Appendix~\ref{suppl:more_ablations}. The source in-distribution statistics $\{\bm\mu_i^S, {\bm\sigma}_i^S\}_{i=0}^N$ are calculated without using the newly inserted prompt.

\textbf{LAME\footnote{\href{https://github.com/fiveai/LAME}{https://github.com/fiveai/LAME}}~\cite{boudiaf2022parameter}.} For fair comparison, we maintain a consistent batch size of 64 for LAME, aligning it with the same batch size used by other methods in our evaluation. We use the kNN affinity matrix with the value of k chosen from \{1, 5, 10, 20\}, and for all experiments, we consistently set it to 5 based on the optimal accuracy observed on ImageNet-C.

\textbf{T3A\footnote{\href{https://github.com/matsuolab/T3A}{https://github.com/matsuolab/T3A}}~\cite{iwasawa2021test}.}  We follow all hyper-parameters that are set in T3A unless it does not provide. Specifically, the batch size is set to 64. The number of supports to restore $M$ is chosen from \{1, 5, 20, 50, 100\}, and for all experiments, we consistently set it to 20 based on the optimal accuracy observed on ImageNet-C.

\textbf{TENT\footnote{\href{https://github.com/DequanWang/tent}{https://github.com/DequanWang/tent}}~\cite{wang2021tent}.} We follow all hyper-parameters that are set in Tent unless it does not provide. Specifically, we use SGD as the update rule, with a momentum of 0.9, batch size of 64 and the learning rate of 0.001. The trainable parameters are all affine parameters of layer normalization layers (except for the experiments in Table~\ref{tab:design_choices}). For updating prompts with SGD optimizer in Table~\ref{tab:design_choices} with entropy loss, the learning rate is set to 0.01 and the number of inserted prompts is set to 3.

\textbf{SAR\footnote{\href{https://github.com/mr-eggplant/SAR}{https://github.com/mr-eggplant/SAR}}~\cite{niu2023towards}.} We follow all hyper-parameters that are set in SAR unless it does not provide. Specifically, we use SGD as the update rule, with a momentum of 0.9, batch size of 64 and the learning rate of 0.001. The entropy threshold $E_0$ is set to 0.4$\times\ln{C}$, where $C$ is the number of task classes. The trainable parameters are the affine parameters of the layer normalization layers from blocks1 to blocks8 for ViT-Base.

\textbf{CoTTA\footnote{\href{https://github.com/qinenergy/cotta}{https://github.com/qinenergy/cotta}}~\cite{wang2022continual}.} We follow all hyperparameters that are set in CoTTA unless it does not provide. Specifically, we use SGD as the update rule, with a momentum of 0.9, batch size of 64 and the learning rate of 0.05. The augmentation threshold $p_{th}$ is set to 0.1. For images below threshold, we conduct 32 augmentations including color jitter, random affine, Gaussian blur, random horizonal flip, and Gaussian noise. The restoration probability of is set to 0.01 and the EMA factor $\alpha$ for teacher update is set to 0.999. The trainable parameters are all the parameters in ViT-Base.

\section{More Ablation Studies}
\label{suppl:more_ablations}

\textbf{Effects of Trade-off Parameter $\lambda$ in Eqn.~(\ref{eq:fitness_function}).}
In the main paper, we simply set the trade-off parameter $\lambda$ in our fitness function (see Eqn.~(\ref{eq:fitness_function})) to 0.4, to balance the magnitude of two terms. Here, we further investigate the sensitivity of $\lambda$, selected from \{0.1, 0.2, 0.3, 0.4, 0.5, 0.6, 0.7, 0.8, 0.9, 1.0\}. From Table~\ref{suppl:tab:lambda ablation}, \methodname maintains comparable accuracy across different values of $\lambda$ as the accuracy variation is not very large, highlighting its insensitivity to changes in $\lambda$. Despite this insensitivity, \methodname achieves better overall performance with $\lambda$ selected from $\{0.3, 0.4, 0.5\}$ in terms of both accuracy and ECE. The observed excellent performance in this range can be attributed to the balanced magnitude of the two terms in Eqn.~(\ref{eq:fitness_function}).

\begin{table*}[h]
    \caption{Sensitivity analyses regarding the trade-off parameter $\lambda$ (see Eqn.~(\ref{eq:fitness_function})) in our \methodname. We report results on ImageNet-C (Gaussian noise, severity level 5) using ViT-Base with batch size 64.}
    \label{suppl:tab:lambda ablation}
\newcommand{\tabincell}[2]{\begin{tabular}{@{}#1@{}}#2\end{tabular}}
 \begin{center}
 \begin{threeparttable}

    \resizebox{0.9\linewidth}{!}{
  \begin{tabular}{lcccccccccc}
  ~ & $\lambda=0.1$ & $\lambda=0.2$ & $\lambda=0.3$ & $\lambda=0.4$ & $\lambda=0.5$ & $\lambda=0.6$ & $\lambda=0.7$ & $\lambda=0.8$ & $\lambda=0.9$ & $\lambda=1.0$\\
  \cmidrule{1-11}

      Acc. (\%, $\uparrow$) & 61.1 & 61.8 & 61.7 & 61.5 & 61.3 & 61.4 & 61.5 & 61.3 & 61.2 & 61.2  \\
      ECE (\%, $\downarrow$) & 5.9 & 3.2 & 2.6 & 2.5 & 2.6 & 2.9 & 3.3 & 3.8 & 4.0 & 4.4  \\

	\end{tabular}
	}
	 \end{threeparttable}
	 \end{center}
  \vspace{-0.1in}
\end{table*}

\begin{table*}[t]
    \caption{Effects of exponential moving average (EMA) (Eqn.~(\ref{eq:testing_mu})) in our Back-to-Source \textit{Activation} (\textit{Act.}) \textit{Shifting} scheme. For \textit{Act. Shifting} w/o EMA, we directly utilize the batch statistics $\bmu_N(\mX_t)$ to calculate the shifting direction $\bd_t$ in Eqn.~(\ref{eq:shifting direction}). We report the average results over 15 corruptions on ImageNet-C (severity level 5) with ViT-Base.}
    \label{suppl:tab:ema statics}
\newcommand{\tabincell}[2]{\begin{tabular}{@{}#1@{}}#2\end{tabular}}
 \begin{center}
 \begin{threeparttable}
 \LARGE
    \resizebox{0.9\linewidth}{!}{
  \begin{tabular}{ll|c|ccccccc}
 	~~~~~~~~~~ & ~ & NoAdapt & $BS=1$ & $BS=2$ & $BS=4$ & $BS=8$ & $BS=16$ & $BS=32$ & $BS=64$  \\
    \cmidrule{1-10}       
        \multirow{2}{*}{Acc. (\%, $\uparrow$)} & \textit{Act. Shifting} w/o EMA & \multirow{2}{*}{55.5} & 0.1  & 56.9  & 58.4  & 58.8  & 59.0  & 59.1  & 59.1\\
       ~ & \textit{Act. Shifting} with EMA (Ours) & ~& 59.0 & 59.1 & 59.1 & 59.1 & 59.1 & 59.1 & 59.2 \\ 
    \cmidrule{1-10}
        \multirow{2}{*}{ECE (\%, $\downarrow$)} & \textit{Act. Shifting} w/o EMA & \multirow{2}{*}{10.5} & 0.0  & 52.4 & 37.5 & 24.5 & 18.0 & 15.1 & 13.8\\
        ~& \textit{Act. Shifting} with EMA (Ours) & ~ & 12.7 & 12.7 & 12.7 & 12.7 & 12.7 & 12.7 & 12.7 \\     
	\end{tabular}
	}
	 \end{threeparttable}
	 \end{center}
  \vspace{-0.1in}
\end{table*}

\begin{table*}[t!]
    \caption{Effects of exponential moving average (EMA) in calculating \textit{Activation Discrepancy} fitness in Eqn.~(\ref{eq:fitness_function}). We replace $\bmu_i(\mX_{t})$ with $\beta*\bmu_i(\mX_{t})+(1-\beta)*\bmu_i(t-1)$ and replace $\bm\sigma_i(\mX_{t})$ in a similar manner. $\beta=1.0$ equals to without using EMA. We report results on ImageNet-C (Gaussian, severity level 5) and ImageNet-R with ViT-Base. }
    \label{suppl:tab:online fitness}
\newcommand{\tabincell}[2]{\begin{tabular}{@{}#1@{}}#2\end{tabular}}
 \begin{center}
 \begin{threeparttable}
 \LARGE
    \resizebox{0.9\linewidth}{!}{
  \begin{tabular}{ll|cccccccccc}
 	~~~~~~~~~~ & ~ & $\beta=0.1$ & $\beta=0.2$ & $\beta=0.3$ & $\beta=0.4$ & $\beta=0.5$ & $\beta=0.6$ & $\beta=0.7$ & $\beta=0.8$ & $\beta=0.9$ & $\beta=1.0$\\
    \cmidrule{1-12}       
        \multirow{2}{*}{ImageNet-C} & Acc. (\%, $\uparrow$)  & 61.0 & 61.4 & 61.5 & 61.4 & 61.5 & 61.7 & 61.7 & 61.2 & 61.9 & 61.5\\
       ~ & ECE (\%, $\downarrow$)   & 9.5 & 7.2 & 4.7 & 4.4 & 3.8 & 3.1 & 3.1 & 3.0 & 3.6 & 2.5 \\ 
    \cmidrule{1-12}
        \multirow{2}{*}{ImageNet-R} & Acc. (\%, $\uparrow$)  & 47.8 & 56.8 & 63.1 & 63.4 & 62.6 & 63.0 & 63.2 & 64.2 & 63.0 & 63.8\\
        ~& ECE (\%, $\downarrow$)  & 19.2 & 8.4 & 2.8 & 2.8 & 3.0 & 2.8 & 2.8 & 2.4 & 3.1 & 2.7 \\     
	\end{tabular}
	}
	 \end{threeparttable}
	 \end{center}
  \vspace{-0.1in}
\end{table*}

\begin{table*}[t]
    \caption{Comparisons with state-of-the-art methods on ImageNet-C (severity level 5) with ViT-Base regarding \textbf{ECE (\%, $\downarrow$)}.  \textbf{BP} is short for \textbf{backward propagation} and the \textbf{bold} number indicates the best result.}
    \label{suppl:tab:imagenet-c-full-precision-ece}
\newcommand{\tabincell}[2]{\begin{tabular}{@{}#1@{}}#2\end{tabular}}
 \begin{center}
 \begin{threeparttable}
 \LARGE
    \resizebox{1.0\linewidth}{!}{
  \begin{tabular}{lcccccccccccccccc>{\columncolor{black!8}}c}
 	\multicolumn{1}{c}{} & \multicolumn{1}{c}{}& \multicolumn{3}{c}{Noise} & \multicolumn{4}{c}{Blur} & \multicolumn{4}{c}{Weather} & \multicolumn{4}{c}{Digital} & \multicolumn{1}{c}{~} \\
        Method & BP & Gauss. & Shot & Impul. & Defoc. & Glass & Motion & Zoom & Snow & Frost & Fog & Brit. & Contr. & Elas. & Pix. & JPEG & Avg. \\
    \cmidrule{1-18}       
        NoAdapt & \ding{55} & 7.5 & 4.6 & 6.6 & 6.5 & 6.2 & 2.6 & 5.0 & 4.7 & 19.7 & 49.2 & 8.6 & 19.3 & 6.0 & 5.0 & 6.2 & 10.5   \\ 
        LAME & \ding{55} & 6.5 & 3.6 & 5.6 & 5.1 & 9.4 & 2.2 & 6.2 & 5.6 & 18.1 & 46.3 & 7.7 & 29.0 & 10.6 & 3.9 & 5.0 & 11.0 \\
        T3A & \ding{55} & 29.6 & 30.0 & 29.6 & 31.1 & 42.0 & 32.1 & 37.1 & 25.7 & 26.2 & 14.7 & 16.6 & 6.1 & 35.0 & 24.4 & 22.5 & 26.8  \\ 
        TENT & \ding{52} & 13.7 & 13.0 & 12.9 & 14.7 & 15.9 & 12.7 & 15.3 & 24.9 & 12.1 & 93.5 & 6.0 & 10.7 & 14.4 & 8.7 & 9.3 & 18.5  \\
        CoTTA & \ding{52} & 4.2 & 2.9 & 4.4 & 7.2 & 12.8 & 7.1 & 11.6 & 4.1 & 0.9 & 5.1 & 3.2 & 15.9 & 8.1 & 5.1 & 5.4 & 6.5\\
        SAR  & \ding{52}    & 7.9 & 7.4 & 7.3 & 9.0 & 9.5 & 7.7 & 9.7 & 6.1 & 6.0 
        & 9.2 & 2.3 & 7.5 & 7.0 & 4.3 & 4.4 & 7.0\\ 
        \cmidrule{1-18}
         \methodname (ours)  & \ding{55}    & 2.5 & 2.4 & 2.5 & 3.4 & 3.3 & 3.0 & 4.0 & 3.2 & 3.3 & 4.8 & 3.0 & 3.4 & 3.2 & 3.0 & 2.8 & \textbf{3.2}\\ 
	\end{tabular}    
	}
	 \end{threeparttable}
	 \end{center}
  \vspace{-0.1in}
\end{table*}

\textbf{Effects of Exponential Moving Average (EMA) in Eqn.~(\ref{eq:testing_mu}).} In Table~\ref{suppl:tab:ema statics}, we investigate the effectiveness of EMA in Eqn.~(\ref{eq:testing_mu}), which is designed to estimate the center of activation features of OOD testing samples accurately. The results show that without Eqn.~(\ref{eq:testing_mu}), \textit{Activation (Act.) Shifting} suffers from notable performance degradation in both accuracy and ECE, particularly when using small batch sizes (\eg, $BS<8$) where batch statistics are less accurate. In contrast, incorporating EMA ensures stable performance even with a batch size of 1, suggesting its effectiveness.

\textbf{Effects of Exponential Moving Average (EMA) in Calculating Eqn.~(\ref{eq:fitness_function})}. In the main paper, we consistently use the batch statistics $\bmu_i(\mX_{t})$ and $\bm\sigma_i(\mX_{t})$ to calculate \textit{Activation Discrepancy} fitness. In Table~\ref{suppl:tab:online fitness}, we further investigate the effects of using EMA to estimate the overall test set statistics for the fitness function calculation. The results indicate a degradation in performance when the balance factor $\beta$ is notably small. This decline is attributed to a biased objective, which encourages batch statistics to compensate for alignment errors in historical overall statistics rather than converging towards the source in-distribution statistics. In contrast, using $\bmu_i(\mX_{t})$ and $\bm\sigma_i(\mX_{t})$ (\ie, $\beta=1.0$) for Eqn.~(\ref{eq:fitness_function}) achieves remarkable performance on both ImageNet-C and ImageNet-R, without requiring additional hyperparameter.

\begin{figure}[t]
\centering
\includegraphics[width=0.8\linewidth]{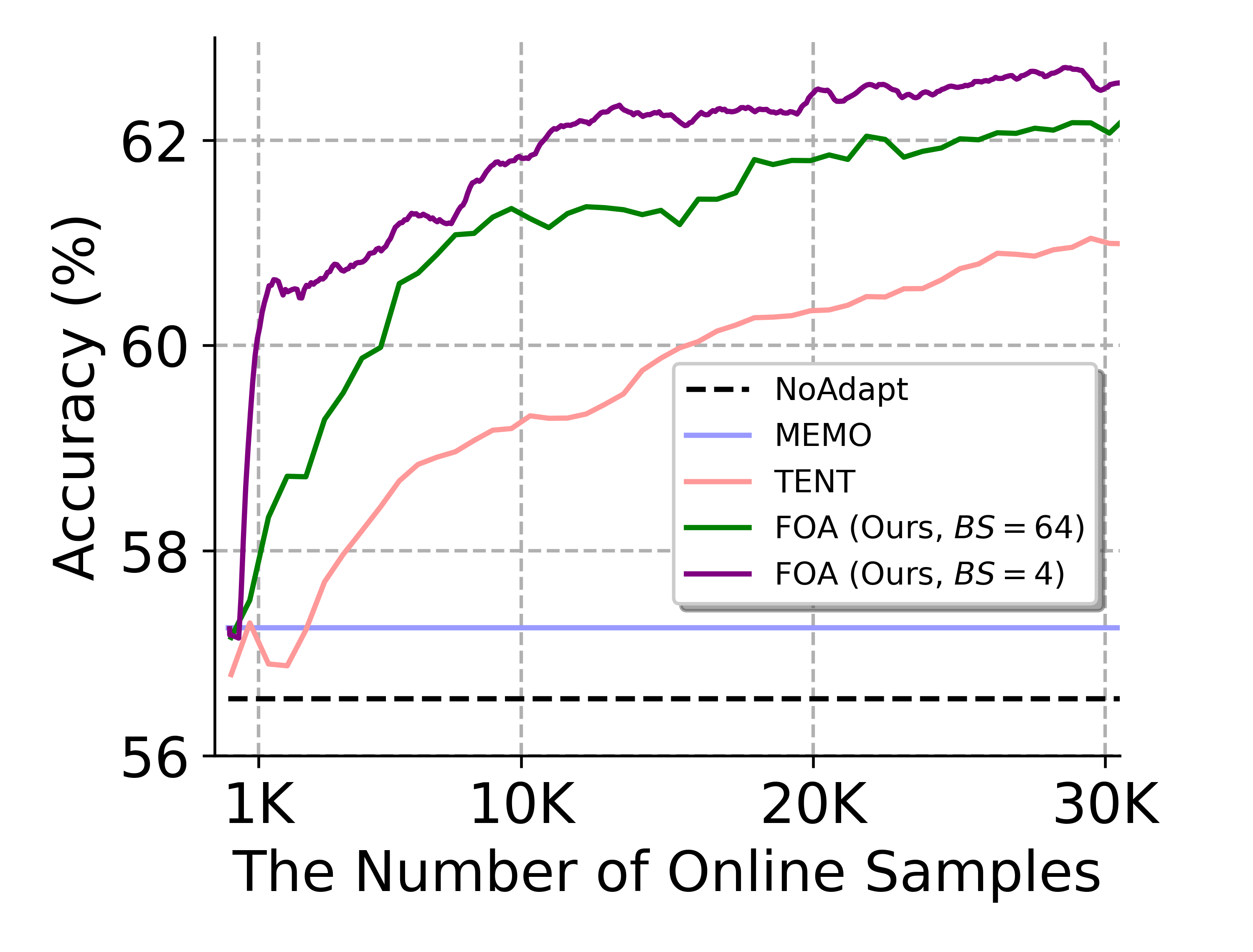}
\vspace{-0.2in}
\caption{Online accuracy comparison with MEMO~\cite{zhang2021memo} on ViT and ImageNet-C (Gaussian noise, severity level 5).}
\label{fig:foa_vs_memo}
\end{figure}

\textbf{Comparison with MEMO~\cite{zhang2021memo} under Different Number of Test Samples.}
To verify the effectiveness of \methodname when only limited test samples are available for adaptation, we record the model's accuracy on the entire test set after the online adaption on $N$ test samples. From Figure~\ref{fig:foa_vs_memo}, our \methodname outperforms MEMO and NoAdapt at the beginning of the test data stream, \ie, less than 200 samples, showing that \methodname does not rely on a long test data stream. The good performance mainly benefits from our Activation Shifting scheme, which boosts the performance a lot in cases of the CMA-based prompt adaptation is inadequate, \eg, at the beginning of adaptation. Here, \methodname with a small batch size (BS) of 4 adapts faster, as a smaller batch size leads to more learning steps.

\section{More Experimental Results}
\label{suppl:more_exps}

\textbf{More Results \wrt ECE.} 
In Tables~\ref{tab:imagenet-c-full-precision} and~\ref{tab:imagenet-rv2asketch-full-precision} of the main paper, we only report the average ECE due to page limits. In this subsection, we provide detailed ECEs for both full-precision and quantized models. From Tables~\ref{suppl:tab:imagenet-c-full-precision-ece} and Table~\ref{suppl:tab:quantized_results-ece}, our \methodname consistently outperforms the state-of-the-art methods among most corruptions and achieves much lower average ECE, \eg, 3.2\% \vs~7.0\% compared with SAR on full precision ViT-Base and 3.8\% \vs~25.9\% compared with T3A on 8-bit quantized ViT-Base. These results demonstrate consistent effects of our \methodname on mitigating the calibration error, further highlighting our effectiveness. Here, the excellent ECEs achieved by \methodname mainly originate from our activation discrepancy regularization item in Eqn.~(\ref{eq:fitness_function}), which alleviates the error accumulation issue of prior gradient-based TTA methods (\eg, TENT~\cite{wang2021tent} and SAR~\cite{niu2023towards}) that may employ imprecise pseudo labels or entropy for test-time model updating.

\begin{table*}[t!]
    \caption{Effectiveness of our \methodname on \textbf{Quantized ViT-Base models}. We report the corruption \textbf{ECE (\%, $\downarrow$)} on ImageNet-C (severity level 5). The \textbf{bold} number indicates the best result.}
    \label{suppl:tab:quantized_results-ece}
\newcommand{\tabincell}[2]{\begin{tabular}{@{}#1@{}}#2\end{tabular}}
 \begin{center}
 \begin{threeparttable}
 \LARGE
    \resizebox{1.0\linewidth}{!}{
  \begin{tabular}{llccccccccccccccc>{\columncolor{black!8}}c}
 	\multicolumn{1}{c}{} & \multicolumn{1}{c}{}& \multicolumn{3}{c}{Noise} & \multicolumn{4}{c}{Blur} & \multicolumn{4}{c}{Weather} & \multicolumn{4}{c}{Digital} & \multicolumn{1}{c}{~} \\
 	 Model & Method & Gauss. & Shot & Impul. & Defoc. & Glass & Motion & Zoom & Snow & Frost & Fog & Brit. & Contr. & Elas. & Pix. & JPEG & Avg. \\
    \cmidrule{1-18}       
        \multirow{3}{*}{8-bit} & NoAdapt  &         8.3 & 5.7 & 7.5 & 7.9 & 4.9 & 3.0 & 4.4 & 5.5 & 20.7 & 50.0 & 10.4 & 14.1 & 5.2 & 6.2 & 8.3 & 10.8  \\  
        & T3A  & 30.0 & 26.0 & 25.7 & 28.7 & 41.9 & 31.2 & 40.1 & 27.2 & 21.6 & 9.6 & 15.8 & 5.7 & 40.7 & 23.1 & 21.0 & 25.9  \\  
 &  \methodname (ours)  & 2.8 & 3.4 & 2.9 & 3.5 & 3.5 & 3.1 & 4.3 & 3.6 & 3.9 & 5.6 & 3.9 & 5.0 & 3.6 & 3.9 & 3.1 & \textbf{3.8} \\   
\cmidrule{1-18}
\multirow{3}{*}{6-bit} &  NoAdapt  & 10.1 & 7.0 & 10.5 & 5.8 & 4.2 & 4.2 & 4.5 & 2.8 & 18.0 & 32.5 & 12.4 & 17.3 & 4.2 & 6.6 & 7.8 & 9.9  \\ 
& T3A     & 31.3 & 27.3 & 26.2 & 43.2 & 50.4 & 35.6 & 49.3 & 33.9 & 22.8 & 14.6 & 16.9 & 7.7 & 44.2 & 26.6 & 22.2 & 30.1 \\ 
 & \methodname (ours)    & 5.6 & 4.9 & 6.7 & 3.1 & 2.4 & 5.0 & 3.5 & 6.6 & 7.1 & 7.0 & 6.0 & 7.7 & 3.3 & 5.8 & 7.7 & \textbf{5.5}\\ 
	\end{tabular}
	}
	 \end{threeparttable}
	 \end{center}
\end{table*}

\end{document}